\documentclass[12pt]{article}

\newcommand{\blind}{0}

\usepackage{amssymb,amsthm,amsthm,amsfonts,latexsym,bbm}
\usepackage{amscd,amsxtra} 
\usepackage{graphics,graphicx,xcolor}
\usepackage{setspace, fancybox, fullpage}
\usepackage{ifpdf}
\usepackage[caption=false]{subfig}
\usepackage{multirow}
\usepackage[colorlinks,linkcolor=blue,citecolor=blue,urlcolor=blue]{hyperref}
\usepackage{colortbl}
\usepackage[comma]{natbib}
\usepackage{comment}
\usepackage{ulem}
\usepackage{authblk} 

\allowdisplaybreaks
\makeatletter
\renewcommand*{\@fnsymbol}[1]{\ensuremath{\ifcase#1\or \dagger\or \ddagger\or
   \mathsection\or \mathparagraph\or \|\or **\or \dagger\dagger
   \or \ddagger\ddagger \else\@ctrerr\fi}}
\makeatother

\theoremstyle{plain}
\newtheorem{theorem}{\sc Theorem}

\theoremstyle{remark}

\theoremstyle{definition}

\newtheorem{algorithm}{\sc Algorithm}

\def\vv{\boldsymbol v}

\def\xv{\boldsymbol x}
\def\yv{\boldsymbol y}
\def\zv{\boldsymbol z}

\def\Xv{\boldsymbol X}

\def\Dbf{\mathbf D}

\newcommand{\muv}{\mbox{\boldmath{$\mu$}}}

\newcommand{\omegav}{\mbox{\boldmath{$\omega$}}}

\def\1v{\mathbf 1}
\def\0v{\mathbf 0}
\def\Id{\mathbf I} 

\newcommand{\R}{\mathbb R}

\newcommand{\E}{\mathbb E}

\def\Pr{\mathrm P}

\newcommand{\var}{\mathop{\rm Var}}

\newcommand{\diag}{\mathop{\rm diag}}

\newcommand{\argmin}{\operatornamewithlimits{argmin}}

\def\half{\frac{1}{2}}

\begin{document}
\def\spacingset#1{\renewcommand{\baselinestretch}%
{#1}\small\normalsize} \spacingset{1}

\if0\blind
{
  \title{\bf Significance Analysis of High-Dimensional, Low-Sample Size Partially Labeled Data}
  \author{Qiyi Lu\thanks{Correspondence to: Qiyi Lu (e-mail: qlu@math.binghamton.edu). Qiyi Lu is a doctoral candidate and Xingye Qiao (e-mail: qiao@math.binghamton.edu) is an Assistant Professor at Department of Mathematical Sciences at Binghamton University, State University of New York, Binghamton, New York, 13902-6000. Qiao's research is partially supported by a collaboration grant from \textit{Simons Foundation} (award number 246649).}\quad and Xingye Qiao\\
    Department of Mathematical Sciences\\
    Binghamton University, State University of New York\\
    Binghamton, New York, 13902-6000\\
    \quad\\
    E-mails:\{qlu,qiao\}{@}math.binghamton.edu\\
    Phone: (607) 777-2147\\
    Fax: (607) 777-2450}
  \maketitle
  \newpage
} \fi

\if1\blind
{
  \bigskip
  \bigskip
  \bigskip
  \begin{center}
    {\LARGE\bf Significance Analysis of High-Dimensional, Low-Sample Size Partially Labeled Data}
  \end{center}
  \medskip
} \fi

\bigskip

\begin{abstract}
Classification and clustering are both important topics in statistical learning. A natural question herein is whether predefined classes are really different from one another, or whether clusters are really there. Specifically, we may be interested in knowing whether the two classes defined by some class labels (when they are provided), or the two clusters tagged by a clustering algorithm (where class labels are not provided), are from the same underlying distribution. Although both are challenging questions for the high-dimensional, low-sample size data, there has been some recent development for both. However, when it is costly to manually place labels on observations, it is often that only a small portion of the class labels is available. In this article, we propose a significance analysis approach for such type of data, namely partially labeled data. Our method makes use of the whole data and tries to test the class difference as if all the labels were observed. Compared to a testing method that ignores the label information, our method provides a greater power, meanwhile, maintaining the size, illustrated by a comprehensive simulation study. Theoretical properties of the proposed method are studied with emphasis on the high-dimensional, low-sample size setting. Our simulated examples help to understand when and how the information extracted from the labeled data can be effective. A real data example further illustrates the usefulness of the proposed method.

\end{abstract}

\vspace{0.15in}

\vfill
\newpage

\spacingset{1.5} 

\setcounter{page}{1}
\abovedisplayskip=8pt
\belowdisplayskip=8pt

\section{Introduction}\label{intro}

Classification and clustering are both important tools in statistical learning. The availability of the class labels distinguish these two main domains. In classification, class labels are provided prior to the analysis, while they are unavailable in the clustering analysis.  A natural statistical question regarding their use is whether classes/clusters are really there. In a setting where binary class labels are observed, we may be interested in testing whether the two classes are from the same distribution. Though often neglected, this is an important step before applying a classification algorithm. In standard statistical textbooks, there are many significance tests, such as two-sample $t$-test, one-way ANOVA, Hotelling's $T^2$ test, and MANOVA. Among these, the two-sample $t$-test and ANOVA are univariate tests. The Hotelling's $T^2$ test and MANOVA are multivariate tests, though both can fail when the dimension $d$ is much greater than the sample size $n$. 

This problem of testing the difference between two classes becomes even more challenging for the high-dimensional, low-sample size (HDLSS) data. The Hotelling's $T^2$ test is very powerful when the dimension is smaller than the sample size. It is invariant under linear transformation. In addition, under the null hypothesis, the distribution of the statistic is known. However, the Hotelling's $T^2$ statistic cannot be computed in the HDLSS setting because the sample covariance matrix is not invertible. There are efforts attempting to overcome this issue, including \citet{dempster1960significance}, \citet{bai1996effect}, \citet{srivastava2008test} and \citet{chen2010two}. These methods use diagonalized versions of the covariance or inverse covariance matrices in the Hotelling's $T^2$ statistic. There are many other treatments, such as \citet{Srivastava2006Multivariate}, \citet{schott2007some} and \citet{srivastava2007multivariate}, which calibrate the distribution of some proposed statistic. In addition, the Direction-Projection-Permutation (DiProPerm) test \citep{wichers2007functional,wei2013direction} has been proved to be very effective for testing the class difference of the HDLSS data.

Besides the difficulty brought from the high dimensionality, in many real problems, it is often the case that there are many observations that are left without class labels (the unlabeled data portion) in a data set. One reason is that it is often difficult or expensive to obtain the class label information, while it may be relatively cheap to obtain the covariate information even for many observations. In such a situation, those aforementioned testing methods which require label information cannot be applied to the whole data set to test the class difference. As a consequence, one may have to forfeit the potentially useful information that resides in the unlabeled data. For instance, many cancer patients are categorized to certain cancer subtypes by radiologists through an inspection of the medical images. However, because of the high health care cost, medical images are easier to obtain than the actual diagnostic. Before a classification algorithm is used to design a data-mining-based early-screening machine \citep[see, for example,][]{Land2012PNN/GRNN,Schaffer2012}, a valuable question to ask is whether the so-called subtypes, many of which may be ad hoc or based on experience, are really there.

One possible, but clearly flawed, solution to this problem is to treat all the data as unlabeled. In the unsupervised context, in the sense that there are no class labels provided for the analysis, clustering algorithms have been broadly applied in many fields. As to determine whether clusters are really there, several methods have been developed to assess the significance of clusters, including \citet{mcshane2002methods, tibshirani2005cluster, suzuki2006pvclust} and \citet{liu2008statistical}. However, these methods are not directly applicable for partially labeled data, unless one forfeits the potentially useful information that resides in the class labels which are available in the labeled data portion of the full data set.

Hence, there seems to be a dilemma in testing partially labeled data: to ignore the unlabeled data completely (and apply a significance test for the labeled data only), versus, to ignore the class labels in the labeled data portion (and use a significance test for clustering). Although each has its own applicability domain, neither looks promising for us. This motivates us to devise a significance testing method for the HDLSS partially labeled data. When class labels are partially provided, the unlabeled data are used to better estimate the sampling distribution. In the meantime, the class labels help to effectively distinguish the two classes even if their difference is small. Our proposed method is named Significance Analysis of HDLSS Partially Labeled Data (SigPal).

\begin{figure}[!ht]\vspace{-1.5ex}
	\begin{center}
	\includegraphics[height=0.8\linewidth]{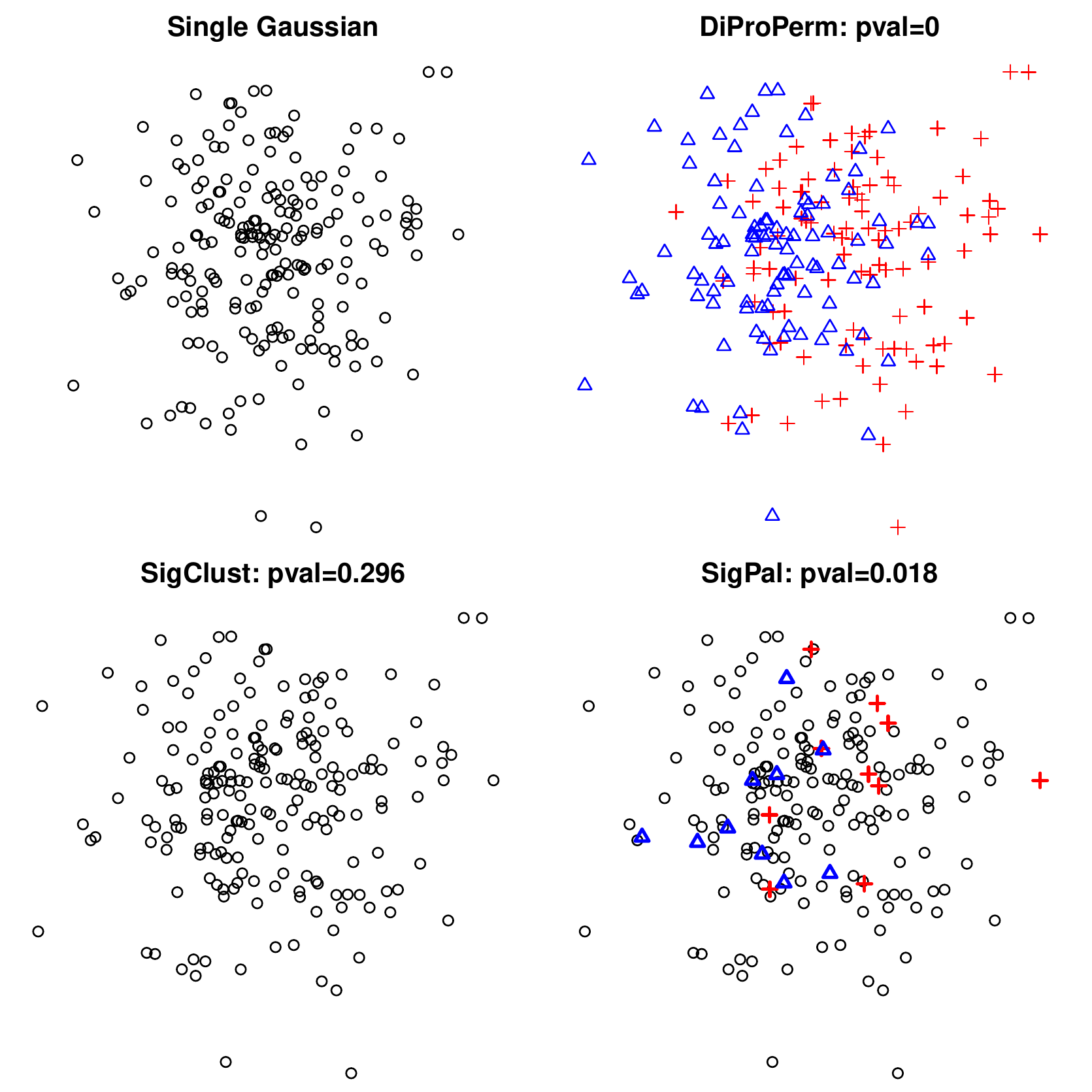}
	\end{center}\vspace{-2ex}
	\caption{The DiProPerm test is applicable when all class labels are known (top-right panel) while SigClust does not require any label information (bottom-left panel). The DiProPerm test correctly concludes that the two classes are indeed from two distributions (with $p$-value$=0$), whereas the SigClust method fails to find this important difference ($p$-value$=0.296$). When the majority of the data are unlabeled with a small portion of labeled data, our proposed SigPal approach can give a significant conclusion with $p$-value$=0.018$, which is close to the DiProPerm result.}
	\label{fig:motivation}
\end{figure}

To illustrate our main idea, we show a toy example under different settings in Figure \ref{fig:motivation}. The data in the top-left panel are generated from a Gaussian distribution and the data in the rest three panels come from a mixture of two Gaussian distributions with a small difference in the mean, $0.5N(-\muv,\Id_2)+0.5N(\muv,\Id_2)$, where $\muv=(0.5,0)'$. To ease the presentation, the toy example is of two-dimensional, though the message applies to the HDLSS data. We show two significance analysis methods for the HDLSS data that inspire our approach, the DiProPerm test of \citet{wichers2007functional} and \citet{wei2013direction}, and the Statistical Significance of Clustering method (SigClust) of \citet{liu2008statistical} and \citet{huang2014statistical}. The DiProPerm test is applicable when all the class labels are known (see the different colors/marker-types in the top-right panel, where each component of the mixture distribution corresponds to one class.) In contrast, SigClust does not require any label information (bottom-left panel). The (empirical) $p$-value of the DiProPerm test turns out to be 0, which leads to a correct conclusion that the two classes are indeed from two distributions, whereas the SigClust method fails to find this important difference ($p$-value$=0.296$). Our proposed SigPal approach is designed for the case shown in the bottom-right panel. Given some labeled data, SigPal can give a significant conclusion with $p$-value$=0.018$, which is close to the DiProPerm result. All these three methods will be introduced or revisited in the next two sections.

The rest of the article is organized as follows. In Section \ref{diproperm_sigclulst}, we review the DiProPerm test and the SigClust method. Section \ref{method} presents our proposed SigPal method. Some theoretical results are studied in Section \ref{theory} which emphasize the HDLSS setting. A comprehensive simulation study and real data case study are provided in Section \ref{numericalstudy}. Section \ref{conclusion} gives some concluding remarks. The appendix is devoted to technical proofs.


\section{DiProPerm Test and SigClust Test}\label{diproperm_sigclulst}
In this section, we review two significance analysis methods, DiProPerm and SigClust. Both methods are specifically designed for testing HDLSS data, although they may be applied to low-dimensional data as well. DiProPerm is used when all the class labels are fully observed while SigClust is applicable when the data set has no class label information. The hypotheses of both methods are slightly different.

\subsection{DiProPerm Test}


In practice, permutation tests are often used for the purpose of testing the class difference, where the null distribution is mimicked by the empirical distribution of the statistic calculated from many randomly permuted data sets. However, for high-dimensional data, some distance measure with direct permutation  may not work. This is because when $d\gg n$, such distance measure will be mainly driven by the error aggregated over dimensions, rather than the true mean difference between classes. To address this issue, a three-step procedure called \textsc{Di}rection-\textsc{Pro}jection-\textsc{Perm}utation test (DiProPerm) was proposed in \citet{wichers2007functional} and further studied in \citet{wei2013direction} for the two-class setting. DiProPerm was designed for data with fully observed labels. It tests the null hypothesis of equality of distributions:

$H_0$: the distributions of the two classes are the same, and 

$H_1$: the distributions of the two classes are not the same. \\
Another item of interest is to test the weaker null hypothesis of equality of means:

$H_0$: the distributions of the two classes have equal means, and

$H_0$: the distributions of the two classes have different means.

\begin{algorithm} The DiProPerm test comprises three steps.
\begin{enumerate}
		\item \textbf{Direction:} a direction which is capable of separating the two classes is found, such as the classification direction from Support Vector Machine \citep[SVM;][]{vapnik1995nature,Cortes1995Support}, Distance Weighted Discrimination \citep[DWD;][]{marron2007distance}, or their hybrids \citep{qiao2015flexible,Qiao2015Distance}.
		\item \textbf{Projection:} all the data vectors are projected to this direction so that a univariate statistic (such as the two-sample $t$-statistic or the mean difference) can be calculated.
		\item \textbf{Permutation:} all the data are randomly relabeled and the first two steps are repeated for $N_{Perm}$ times ($N_{Perm}$ may be 1000.) An empirical $p$-value is calculated to assess the statistical significance (the proportion of the statistics from the permutation set that are greater than that from the data).
\end{enumerate}
\end{algorithm}

\begin{figure}[!ht]\vspace{-1.5ex}
	\begin{center}
	\includegraphics[height=0.4\linewidth]{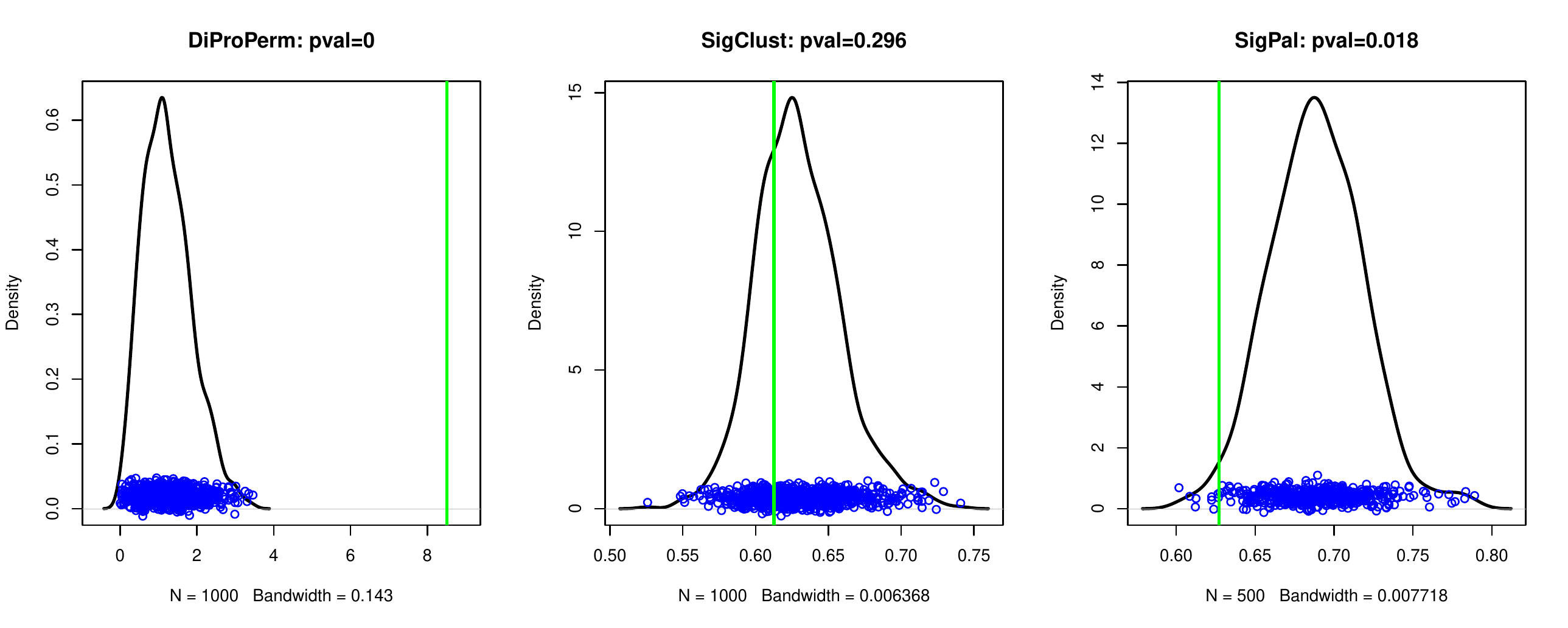}
	\end{center}\vspace{-2ex}
	\caption{Illustration of the calculation of the $p$-values for DiProPerm (left), SigClust (middle) and SigPal (right). The test statistics for the permutation/simulation set are shown as blue jitter points, while those for the original data are shown as the green vertical lines. The empirical $p$-value for DiProPerm is the proportion of the statistics from the permutation set that are greater than that from the data, which is 0 in this case. The empirical $p$-values for both SigClust and SigPal are the proportions of the statistics from the simulation set that are less than that from the data, which are 0.296 and 0.018 respectively. }
	\label{fig:pval}
\end{figure}

In Figure \ref{fig:pval}, we illustrate how the $p$-value in DiProPerm is calculated, using the same data as shown in Figure \ref{fig:motivation}. In the left plot, we perform a DiProPerm test with 1000 permutations. The test statistics calculated for the permutations are shown as the blue jitter points, while that for the original data is the green vertical line. Here the mean difference is chosen as the statistic. The greater the statistic is, the more significantly different the two classes are. Hence, the empirical $p$-value is calculated as the proportion of the statistics from the permutation set that are greater than that from the data, which is 0 in this case.

The main idea of DiProPerm is to measure the difference between two high-dimensional data subsets by the difference between their 1-dimensional projections onto some appropriate direction. DiProPerm is a powerful test in many settings and it is a nonparametric procedure that does not have many assumptions. Note that here class labels are required to find the projection direction (by a classification method such as SVM or DWD) and to calculate the test statistic ($t$-statistic or mean difference).


\subsection{SigClust}

SigClust, proposed by \citet{liu2008statistical} and improved by \citet{huang2014statistical}, is a clustering evaluation tool for the HDLSS data which aims to answer the question whether clusters are really there. That is, it has the following hypotheses:

$H_0$: the data are from a single Gaussian distribution, and

$H_1$: the data are not from a single Gaussian distribution.

There is no reference to the notion of class in the hypotheses above. SigClust is based on the vision of cluster as a subset of the data that can be reasonably modeled as coming from a single Gaussian distribution (with some covariance matrix). The Gaussian assumption has been previously used by \citet{sarle1993modeclus} and \citet{mclachlan2004finite}. \citet{huang2014statistical} mentioned that this assumption may lead to some important consequences. For example, it is possible that none of Cauchy, Uniform, or even $t$ distributed data may be viewed as a single cluster in this sense. While it may seem to be a strong assumption, it is reasonable in the challenging HDLSS situation because it allows real HDLSS data analysis with wide use in bioinformatics applications \citep{chandriani2009core,verhaak2010integrated}.

Assume that a data set $\{\xv_i,~i=1,\ldots,n\}$ is obtained from an unknown Gaussian distribution with covariance $\Sigma$, where $\xv_i\in\R^d$ is the observed covariates. The idea of SigClust is to approximate the null distribution of a test statistic by simulating from a single Gaussian distribution that fits to the data. The $p$-value in SigClust is taken to be the lower quantile of the null distribution, defined by the test statistic from the original data. It is similar to the DiProPerm test except that it  performs simulation instead of permutation and it relies on a multivariate statistic instead of a univariate statistic after projection.  

Specifically, the 2-means Cluster Index (CI) is used as the test statistic. It is defined as the sum of the within-cluster sums of squares about the cluster means, divided by the sum of squares about the overall mean,
$$CI=\frac{\sum_{k=1}^2\sum_{j\in C_k}\|\xv_j-\bar\xv^{(k)}\|^2}{\sum_{j=1}^n\|\xv_j-\bar\xv\|^2},$$
where $C_k$ denotes the sample index set of the $k$th cluster and $\bar\xv^{(k)}$ represents the mean of the $k$th cluster, for $k = 1, 2$. The smaller the CI, the larger the proportion of the overall variation that is explained by the clusters. Note that no predefined class labels are needed when computing the CI, as the cluster assignment $C_k$ is obtained concurrently by a clustering algorithm.

Here the simulation from the null distribution is a critical part. As CI is location and rotation invariant, it is enough to work only with a Gaussian null distribution with a mean at the origin and a diagonal covariance matrix $\Lambda$. Hence, an essential part of the SigClust test is the estimation of the eigenvalues of the covariance matrix $\Sigma$.

\begin{algorithm} The SigClust procedure is summarized as follows.

\begin{enumerate}
		\item \textbf{Initialization:} obtain a two-cluster assignment $(k=2)$ from an application of a clustering algorithm, such as $k$-means. The CI is then calculated for the original data set based on the cluster assignment.
		
		\item \textbf{Simulation:} simulate data from the null distribution: $(X_1,\cdots,X_d)$ are independent with $X_j\sim \textsl{N}(0,\hat\lambda_j)$, where $(\hat\lambda_1,\cdots,\hat\lambda_d)$ is an estimate of the eigenvalues $(\lambda_1,\cdots,\lambda_d)$ of the covariance matrix $\Sigma$. Then calculate the corresponding CI on each simulated data after performing clustering in the same manner as in the \textbf{Initialization} step.
		
		\item \textbf{Testing:} repeat the \textbf{Simulation} step for $N_{Sim}$ times to obtain an empirical distribution of CI based on the null hypothesis ($N_{Sim}$ may be 1000). Then calculate the empirical $p$-value to assess the statistical significance (the proportion of the CI from the simulation set that are less than the CI from the original data.)
		
\end{enumerate}
\end{algorithm}

The middle plot in Figure \ref{fig:pval} illustrates how the $p$-value in SigClust is calculated. Similar to the DiProPerm case (left plot), the blue jitter plots are the statistics from the simulations, and the green vertical line is that for the original data. Recall that the smaller the statistic (chosen as the CI) is, the more significantly different the two clusters are. Hence the empirical $p$-value is the proportion of the CI's from the simulation set that are less than that from the original data, which is 0.296 in this case.

The covariance estimation in the \textbf{Simulation} step can be challenging, especially when the data are HDLSS. Although we only need to estimate the eigenvalues of the covariance matrix, which greatly reduces the number of parameters to be estimated, this problem is still not trivial in the HDLSS setting. \citet{liu2008statistical} used a hard-thresholding approach for eigenvalue estimation. In particular, they first estimate the background noise level using a robust variance estimate. Then those estimated eigenvalues smaller than the background noise level are replaced with the noise level, that is,
\[
    \hat\lambda_j=\left\{
                \begin{array}{ll}
                  \tilde{\lambda_j} \quad &\textit{if~} \tilde{\lambda_j}\geq \hat{\sigma}_N^2\\
                  \hat{\sigma}_N^2  \quad &\textit{if~} \tilde{\lambda_j} < \hat{\sigma}_N^2
                \end{array}
              \right.,
\]
where $(\tilde{\lambda}_1,\cdots,\tilde{\lambda}_d)$ are the eigenvalues of the sample covariance matrix and $\hat{\sigma}_N^2$ is the estimated background noise level. 

\citet{huang2014statistical} later showed that with eigenvalue estimation using hard-thresholding, SigClust can be seriously anti-conservative if the first eigenvalue is relatively large. They proposed a less-aggressive soft-thresholding variant which greatly improved the performance of SigClust. Specifically, they use
\[
    \hat\lambda_j=\left\{
                \begin{array}{ll}
                  \tilde{\lambda_j}-\tau \quad &\textit{if~} \tilde{\lambda_j}\geq \tau+\hat{\sigma}_N^2\\
                  \hat{\sigma}_N^2  \quad &\textit{if~} \tilde{\lambda_j} < \tau+\hat{\sigma}_N^2
                \end{array}
              \right..
\]
A detailed definition of $\tau$ can be found in \citet{huang2014statistical}.

%
%
%

\section{Significance Analysis for Partially Labeled Data} \label{method}
In this section, we first state the background and hypotheses of our problem, followed by a presentation of our proposed method.

\subsection{Background and Hypotheses}

Consider a binary testing problem for a data set with the labeled data portion $\{(\xv_i,y_i),~i=1,...n_l\}$, and the unlabeled data portion $\{\xv_{n_l+j},~ j=1,\cdots,n_u\}$. All the $\xv_i$'s and $\xv_{n_l+j}$'s are $d$-dimensional covariates and the class label $y_i\in \{-1,1\}$. The total sample size is  $n=n_l+n_u$. Let $\theta=n_l/n$ be the proportion of the labeled data, and $1-\theta$ is the proportion of the unlabeled data. We formulate our proposed SigPal procedure as a hypothesis testing problem with the following hypotheses:

$H_0$: the data come from a single Gaussian distribution, and

$H_1$: the conditional distributions of the two classes are different and hence the data are not from a single Gaussian distribution.

It is worth comparing our alternative hypothesis with those of the DiProPerm and SigClust tests. Since not every class label is observed, the notion of the class, as in the alternative hypothesis of DiProPerm, is moot or murky. Technically, our alternative hypothesis is neither an intersection nor a union of the previous alternative hypotheses. In the framework of SigPal, there exists an underlying class label for each observation. We are interested in the difference in the conditional distributions of the data with respect to this underlying label. Our goal is to infer the significance of the otherwise fully observed data based on the partially labeled data with the help of the covariate information of the whole data. Lastly, the fact that the conditional distributions are different implies that the data are not from a single Gaussian distribution (but the converse is not true.)

\subsection{Proposed Method}

With different values of the proportion of the labeled data, $\theta$, we may consider different ways to address the problem. When $\theta$ is close to 1, which means the majority of the data have label information available, then one may just ignore the small amount of unlabeled data and perform a DiProPerm test on the labeled data portion only. When $\theta$ is very close to 0, which means the majority of the data are unlabeled, then one can simply apply SigClust regardless of the few class labels. While we may lose some useful information from the data, such simplifications effectively reduce the complexity of the problem. In this article, we are more interested in the case when $\theta$ is not close to 0 or 1. We propose a Significance Analysis for Partially Labeled Data (SigPal) which makes use of the extra label information, compared to the SigClust procedure.

\begin{algorithm} The SigPal procedure consists of the following three steps. 

\begin{enumerate}
			\item \textbf{Initialization:} obtain the predicted class assignments for the unlabeled data by applying a semi-supervised classification/clustering method to the full data set. A test statistic is then calculated for the original data based on both observed and predicted class labels (our choice is CI in this article).
			
			\item \textbf{Simulation:} simulate data from the null distribution: $(X_1,\cdots,X_d)$ are independent with $X_j\sim \textsl{N}(0,\hat\lambda_j)$, where $(\hat\lambda_1,\cdots,\hat\lambda_d)$ is an estimate of the eigenvalues $(\lambda_1,\cdots,\lambda_d)$ of the covariance matrix $\Sigma$. Randomly place class labels to $n_l$ observations in the simulated set and then calculate the corresponding test statistic after performing semi-supervised classification/clustering in the same manner as in the \textbf{Initialization} step.
							
			\item \textbf{Testing:} repeat the \textbf{Simulation} step for $N_{Sim}$ times to obtain an empirical distribution of the test statistic based on the null hypothesis. Calculate the empirical $p$-value (the proportion of the CI's from the simulated data that are less than that from the original data) to assess the statistical significance.
\end{enumerate}
		
\end{algorithm}



The right plot in Figure \ref{fig:pval} shows the $p$-value calulation for SigPal. Similar to the SigClust, the empirical $p$-value is the proportion of the CI's from the simulation set that are less than that from the original data, which is 0.018 in this case.

To calculate the statistic in the  \textbf{Initialization} step, we need to assign labels for the unlabeled portion. This is similar to the application of a clustering algorithm in SigClust. Such label assignment can be done either by modifying a classification method or by modifying a clustering algorithm.

\begin{itemize}
	\item While we could simply use a classifier trained from the labeled portion to predict the class label for the unlabeled portion, a semi-supervised classification method is more reasonable here since it takes the covariate information in the large number of unlabeled observations into account.  Possible choices of the semi-supervised classification method include Semi-Supervised Sparse Linear Discriminant Analysis \citep[$S^3$LDA;][]{lu2015sparse}, transductive SVM \citep[TSVM;~][]{vapnik1998statistical, chapelle2006semi, wang2007transductive}, the large-margin based methods \citep{wang2007large, wang2009efficient} and the bootstrap method \citep{collins1999unsupervised}. $S^3$LDA combines the classical linear discriminant analysis and a machine learning oriented technique, and takes advantage of the unlabeled data to boost the classification performance. 
	\item Similarly, though we could just run a clustering algorithm for the whole data set to assign labels, we would like to borrow the strength in the labeled data portion. A semi-supervised clustering algorithm, which identifies clusters with constraints imposed by known labels, would be more appropriate in this case. Possible semi-supervised clustering algorithms include constrained $k$-means \citep[COP-KMEANS;][]{wagstaff2001constrained}, and others. COP-KMEANS allows a must-link constraint which specifies that certain instances have to be placed in the same cluster.
\end{itemize}

Once the class/cluster labels are assigned, a test statistic can be calculated. Options include the Hotelling's $T^2$ statistic, the CI, and some one-dimensional statistics (such as two-sample $t$-statistic or mean difference) after projections as in DiProPerm. CI is more favorable here since it is location and rotation invariant and can be efficiently computed. It also facilitates the comparison between SigPal and SigClust in our numerical studies. It can be shown that for certain low-dimensional examples, the CI is equivalent to the two-sample $t$-statistic.

Similar to SigClust, we use simulation in lieu of permutation, and make use of the soft-thresholding method \citep{huang2014statistical} for eigenvalue estimation. SigPal randomly labels some observations in the simulated data. This extra step is essential to mimic the true null distribution of the test statistic.



As will be shown later, although SigClust has some power when the signal within the data is relatively large, it is substantially less powerful when the signal is weak. SigPal, on the other hand, has a great power in both cases. Secondly, when the data come from a mixture of two Gaussian distributions and the mean difference is large enough, the labeled data may not provide additional boost in the power compared to SigClust. In this case, it may make sense to simply apply SigClust and ignore the label information. As will be seen in the later sections, the strength of SigPal lies on the usefulness of the labeled data: it is visibly more powerful than SigClust when  the signal is small.


\section{Theoretical Property} \label{theory}

In this section, we provide some theoretical justification for the SigPal method. We first derive the relationship between the theoretical version of the CI (TCI) and the eigenvalues of the covariance. Specifically, we assume that $\Xv\sim \textsl{N}(\0v,\Sigma)$ and consider using $S^3$LDA for class assignment. The theoretical $S^3$LDA \citep{lu2015sparse} coefficient $\hat{\omegav}$ is defined as 
\begin{align*}
	\hat{\omegav} &=\argmin_{\|\omegav\|=1,b=0}\quad\E_{(\Xv,Y)}(Y-(\omegav'\Xv+b))^2+C\E_{\Xv}(1-|\omegav'\Xv+b|)_+, \\
	&=\argmin_{\|\omegav\|=1}\quad\E_{(\Xv,Y)}(Y-\omegav'\Xv)^2+C\E_{\Xv}(1-|\omegav'\Xv|)_+,
\end{align*}
where $C>0$ is a constant. We consider the case where the effect of the unlabeled data portion dominates, that is, we let $C\rightarrow \infty$. The relationship between the theoretical cluster index (TCI) for SigPal and the eigenvalues of $\Sigma$ is stated in Theorem \ref{theorem_tci}.

\begin{theorem}
Suppose that $\Xv\sim \textsl{N}(\0v,\Sigma)$, $\Pr(Y=+1)=\Pr(Y=-1)=1/2$ and the proportion of the labeled data is $\theta$.  Assume that $\Sigma$ has an eigen-decomposition $\Sigma=V'\Lambda V$, where $\Lambda=\diag(\lambda_1,\cdots,\lambda_d)$ with $\lambda_1\geq\lambda_2\geq\cdots\geq\lambda_d$. Let $\vv_1$ be the first principal component direction. Then when $C\rightarrow\infty$, $\hat{\omegav}=\vv_1$, and the corresponding TCI is 
$$TCI=1+\theta-\frac{2}{\pi}(1-\theta)^3\frac{\lambda_1}{\sum_{j=1}^d\lambda_j}.$$
\label{theorem_tci}
\end{theorem}

Theorem \ref{theorem_tci} shows that given $\theta$, the optimal TCI only relies on the largest eigenvalue $\lambda_1$ and the sum of eigenvalues $\sum_{i=1}^d\lambda_i$. In practice, the estimations of these two quantities have a critical impact on the $p$-values, defined as the proportion of the CI's from the simulated data that are less than that from the original data. In particular, let $\hat \lambda_i$ denote the estimate of $\lambda_i$. Assume that $\theta=1/2$ (for a mere illustration), then the difference between the true TCI (the one for the original distribution) and the TCI resulting from a Gaussian distribution with covariance $\hat\Lambda$ (the estimated $\Lambda$) is proportional to,
\begin{equation}\label{eq:diff}
E=\frac{\hat\lambda_1}{\sum_{i=1}^d\hat\lambda_i}-\frac{\lambda_1}{\sum_{i=1}^d\lambda_i}.
\end{equation}
For hard-thresholding method, define the potential biases in the estimation of $\lambda_1$ and $\sum_{i=1}^d\lambda_i$ as $\delta_1$ and $\Delta$ respectively. Then
$$E=\frac{\lambda_1+\delta_1}{\sum_{i=1}^d\lambda_i+\Delta}-\frac{\lambda_1}{\sum_{i=1}^d\lambda_i}=\frac{\sum_{i=1}^d\lambda_i\delta_1-\lambda_1\Delta}{\sum_{i=1}^d\lambda_i(\sum_{i=1}^d\lambda_i+\Delta)}.$$
The hard-thresholding method will tend to be anti-conservative when $E< 0$, or $\lambda_1\Delta>\delta_1\sum_{i=1}^d\lambda_i$, that is, when the first eigenvalue is large relative to the rest (assuming that $\Delta$ is positive, which is very likely for the hard-thresholding method since the smallest eigenvalues are replaced by the background noise level.) On the other hand, the soft-thresholding method is energy preserving in the sense that the sum of the soft eigenvalues is the sum of the sample eigenvalues, and thus $\Delta=0$. It follows that when $\hat\lambda_1<\lambda_1$, that is, when the the largest eigenvalue is under-estimated, the soft-thresholding method will be anti-conservative. This happens when the first eigenvalue is only a little larger than the background noise. A detailed discussion about the results of hard-thresholding and soft-thresholding methods can be found in \citet{huang2014statistical}.

We further explore the impact of the estimation of $\lambda_i$ on the TCI in SigPal and SigClust. Let $\textit{TCI}_{\textit{SigPal}}$ and $\textit{TCI}_{\textit{SigClust}}$ denote the TCI's of SigPal and SigClust respectively, and $\hat{\textit{TCI}}_{\textit{SigPal}}$ and $\hat{\textit{TCI}}_{\textit{SigClust}}$ the TCI's from Gaussian distributions based on estimated covariance (that is, the simulated data in both SigPal and SigClust). By Theorem 3 of \citet{huang2014statistical}, $$\textit{TCI}_{\textit{SigClust}}=1-\frac{2}{\pi}\frac{\lambda_1}{\sum_{i=1}^d\lambda_i}.$$ Then when $\theta=1/2$ (again, for an illustration), the differences are
\begin{align}
\textit{TCI}_{\textit{SigPal}}-\textit{TCI}_{\textit{SigClust}}=1/2+\frac{7}{4\pi}\frac{\lambda_1}{\sum_{i=1}^d\lambda_i}.\label{eq:tci_diff}\\
\hat{\textit{TCI}}_{\textit{SigPal}}-\hat{\textit{TCI}}_{\textit{SigClust}}=1/2+\frac{7}{4\pi}\frac{\hat\lambda_1}{\sum_{i=1}^d\hat\lambda_i}.\label{eq:tci_diff2}
\end{align}
A SigPal/SigClust test gains power if $\textrm{TCI}\ll\hat{\textrm{TCI}}$, that is, the TCI for the original data is less than the TCI from simulated data. When $\frac{\hat\lambda_1}{\sum_{i=1}^d\hat\lambda_i}>\frac{\lambda_1}{\sum_{i=1}^d\lambda_i}$, $\hat{TCI}_{SigPal}-TCI_{SigPal}$ is greater than $\hat{TCI}_{SigClust}-TCI_{SigClust}$ due to (\ref{eq:tci_diff}) and (\ref{eq:tci_diff2}). Therefore, SigPal is more powerful than SigClust when $\frac{\hat\lambda_1}{\sum_{i=1}^d\hat\lambda_i}>\frac{\lambda_1}{\sum_{i=1}^d\lambda_i}$. Particularly for soft-thresholding method, $\sum_{i=1}^d\hat\lambda_i=\sum_{i=1}^d\lambda_i$ because it is energy preserving. When the first eigenvalue is very large relative to the others, it is easier to be over-estimated, in which case SigPal will be more powerful.

\begin{figure}[!t]\vspace{-1.5ex}
	\begin{center}
	\includegraphics[height=0.5\linewidth]{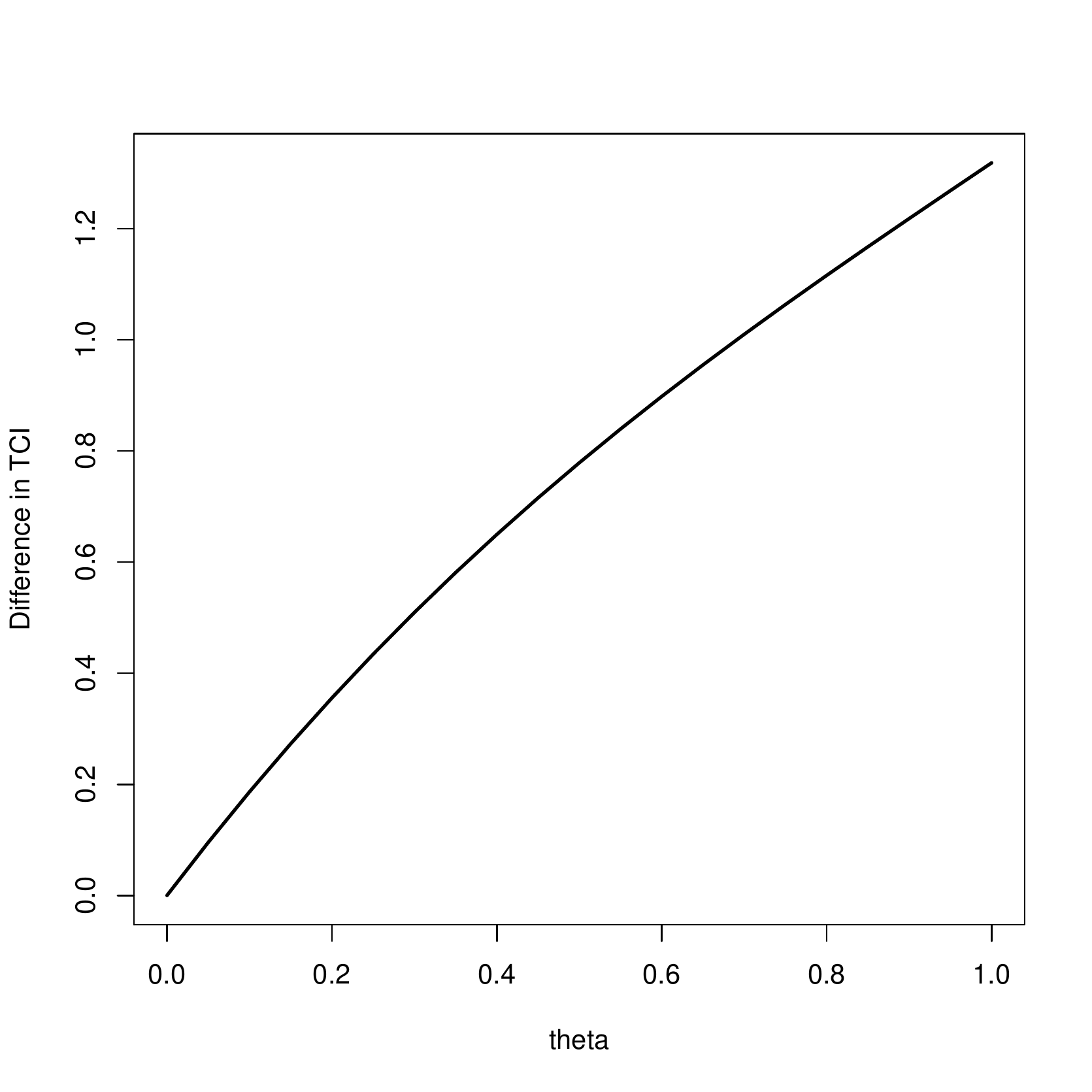}
	\end{center}\vspace{-2ex}
	\caption{The difference $\textit{TCI}_{\textit{SigPal}}-\textit{TCI}_{\textit{SigClust}}$ with respect to $\theta$, the proportion of the labeled data.}
	\label{fig:tci_diff}
\end{figure}

It is also of interest to look at the change of the difference $\textit{TCI}_{\textit{SigPal}}-\textit{TCI}_{\textit{SigClust}}$ with respect to $\theta$ when the eigenvalues are fixed. Assume that $\frac{\lambda_1}{\sum_{i=1}^d\lambda_i}=1/2$, then we have the difference
$$\textit{TCI}_{\textit{SigPal}}-\textit{TCI}_{\textit{SigClust}}=\frac{1}{\pi}[\theta^3-3\theta^2+(\pi+3)\theta].$$
We plot the difference in Figure \ref{fig:tci_diff}, which shows that the difference is almost linear in $\theta$.  It also shows that when $\theta=0$, the difference equals 0 too. This is because when the proportion of the labeled data is 0, then all of the data are unlabeled, in which case the data set is reduced to the SigClust setting. Hence, $\textit{TCI}_{\textit{SigPal}}=\textit{TCI}_{\textit{SigClust}}$. As $\theta$ increases, labeled data play a more and more important role, and the difference of $\textit{TCI}_{\textit{SigPal}}$ and $\textit{TCI}_{\textit{SigClust}}$ increases almost linearly with respect to $\theta$.

In the next theorem, we study some asymptotic properties of SigPal. Since our main interest is in the HDLSS data, we choose to consider asymptotics for $d\rightarrow\infty$ with $n$ fixed. Such kind of HDLSS asymptotic properties  were previously studied by \citet{Hall2005Geometric, Ahn2007high, liu2008statistical, jung2009pca, qiao2010weighted, jung2012boundary, qiao2015flexible}, among others.

Consider a mixture of two Gaussian distributions, $\eta N(\0v,\Dbf)+(1-\eta)N(\muv,\Dbf)$, where $\eta\in(0,1)$ is the proportion for the mixture, $\muv=(a_1,\cdots,a_d)'$ a constant vector and $\Dbf$ is a diagonal matrix with diagonal elements $\lambda_1\geq\lambda_2\geq\cdots\geq\lambda_d$. Note that the theoretical variance for the $i$th variable of the mixture is $\lambda_i+\eta(1-\eta)a_i^2$. We assume that $\lambda_1$ and $a_i$'s are bounded.

\begin{theorem}
Suppose that the data come from a mixture of two Gaussian distributions, $\eta N(\0v,\Dbf)+(1-\eta)N(\muv,\Dbf)$, where $\eta\in(0,1)$, $\muv=(a_1,\cdots,a_d)'$ and $\Dbf$ is a diagonal matrix with diagonal elements $\lambda_1\geq\lambda_2\geq\cdots\geq\lambda_d$. Let $n_1$ and $n_2$ be the sample sizes with $min(n_1,n_2)>0$ and $n_1+n_2=n\geq 3$. Assume that $\sum_{j=1}^d\lambda_j=O(d^{\beta})$ with $0\leq \beta<1$, $\sum_{j=1}^d a_j^2=O(d)$, $\sum_{j=1}^d a_j^2\lambda_j=O(d^{\gamma})$ with $\gamma< 2$ and $\max_j(\lambda_j+\eta(1-\eta)a_j^2)\leq M$ with $M>0$ a fixed constant. If $\Dbf$ is known, then for a fixed $n$, the corresponding SigPal $p$-value converges to 0 in probability as $d\rightarrow \infty$.
\label{theorem_asymp}
\end{theorem}

\citet{liu2008statistical} studied a similar result for SigClust in a special case  when $a_1=a_2=\cdots=a_d=a$, where $a$ is a fixed constant. While we add some more assumptions, the theorem shows the asymptotic property for a more general setting. Theorem \ref{theorem_asymp} shows that if the data come from a mixture of two Gaussian distributions, then under some assumptions, SigPal tends to reject the null hypothesis when $n$ is fixed and $d\rightarrow \infty$. This result justifies the usefulness of SigPal in the HDLSS data setting.

\section{Numerical Study} \label{numericalstudy}
In this section we use simulated and real examples to demonstrate the effectiveness of our proposed method. 

\subsection{Simulations}

We use the same simulation setting as in \citet{liu2008statistical} and \citet{huang2014statistical}. Three types of examples are studied, including three cases under both the null and alternative hypotheses. The sample size for all examples is $n=40$ and dimension $d=300$. In the first case, we consider examples of data under the null hypothesis, that is, having one cluster generated from a single Gaussian distribution. In each example, we check the type-I error by studying how often SigPal incorrectly rejects the null hypothesis $H_0$. In the second and the third cases, we consider data from a collection of mixtures of two Gaussian distributions with different signal sizes and explore the power of our method in terms of how often it correctly rejects the null hypothesis. For simplicity, we consider diagonal covariance matrix $\Dbf$ because of the rotation invariance property of CI.

For each simulation, we consider two class assignment methods to be applied in SigPal, namely, $S^3$LDA and COP-KMEANS. They are applied for both original data and simulated data before we calculate CI. Under the null hypothesis, theoretically the $p$-value should follow the Uniform [0,1] distribution. Then a level $\alpha$ test rejects the null hypothesis with probability $\alpha$ when $\Dbf$ is known. This can be shown by a direct use of the standard probability integral transformation theorem. To simplify the computation, we fix the tuning parameters in $S^3$LDA (the goal here is not perfect classification.) We also consider an option which uses $L_1$-LDA for labeled data only on the simulated data while still using $S^3$LDA on the original data. Note that this does not affect the size of the test, however, could sacrifice the power. We apply these different versions of SigPal to compare with SigClust in each case. To make the notation simple, we use $L_1$-LDA to denote SigPal with $S^3$LDA on the original data and $L_1$-LDA on the simulated data, and use $S^3$LDA and COP-KMEANS to denote SigPal with corresponding methods on both original and simulated data.

\subsubsection{Case 1: One Cluster}

Suppose that the data are generated from a single multivariate Gaussian distribution with covariance $\Dbf$, where $\Dbf$ is diagonal with diagonal elements $(\underbrace{v,\cdots,v}_{w},1,\cdots,1)$, that is, there are $w$ $v$'s and $(d-w)$ 1's. We randomly select 20 observations to be labeled from all the 40 observations and consider 14 combinations of $(v,w)$ as shown in Table \ref{one_cluster}. Each simulation is repeated 100 times.

\begin{table}[!t]
\centering
\begin{tabular}{c|c||c|c|c||c}
\hline
v    &    w & $L_1$-LDA & $S^3$LDA     & COP-KMEANS & SigClust  \\\hline
100  &    1 &       0      &        3		  &     4       &    0         \\
 50  &    2 &       1      &        4		  &     3       &    0         \\
 20  &    5 &       0      &        1		  &     1       &    0         \\
 10  &   10 &       0      &        2		  &     2       &    0         \\
  1  &    1 &       0      &        0		  &     0       &    0         \\
	3  &    1 &       0      &        0		  &     0       &    0         \\
	5  &    1 &       0      &        0		  &     0       &    0         \\
 10  &    1 &       0      &        0		  &     2       &    0         \\
 20  &    1 &       0      &        2		  &     4       &    0         \\
 50  &    1 &       1      &        3 	  &     4       &    0         \\
  1  &    5 &       0      &        0		  &     0       &    0         \\
 10  &    5 &       0      &        1		  &     2       &    0         \\
 20  &    5 &       0      &        1		  &     2       &    0         \\
 50  &    5 &       0      &        2		  &     1       &    0         \\\hline
\end{tabular}
\caption{Summary table for the one cluster case over 100 replications based on different methods under different settings (different pairs of $v$ and $w$). The numbers of empirical $p$-values which are less than 0.05 are reported.
}\label{one_cluster}
\end{table}

In Table \ref{one_cluster}, as expected, all methods maintain the size (fewer than 5\% of the $p$-values are less than 0.05.) $L_1$-LDA is very conservative for all the 14 settings as it almost never rejects the null. A possible explanation is that under the null hypothesis, that is, when the data are generated from a single Gaussian distribution, a semi-supervised classification method like $S^3$LDA makes the class difference even smaller than applying $L_1$-LDA on labeled data only (since the former attempts to incorporate the useless information.) As a consequence, the CI from the original data (after applying $S^3$-LDA) is often greater than the CI's from the simulated data (after applying $L_1$-LDA), and hence the $p$-value is often very large.

\subsubsection{Case 2: Mixture of Two Gaussian Distributions with Signal in One Coordinate Direction}

\begin{figure}[!t]\vspace{-1.5ex}
	\begin{center}
	\includegraphics[height=0.8\linewidth]{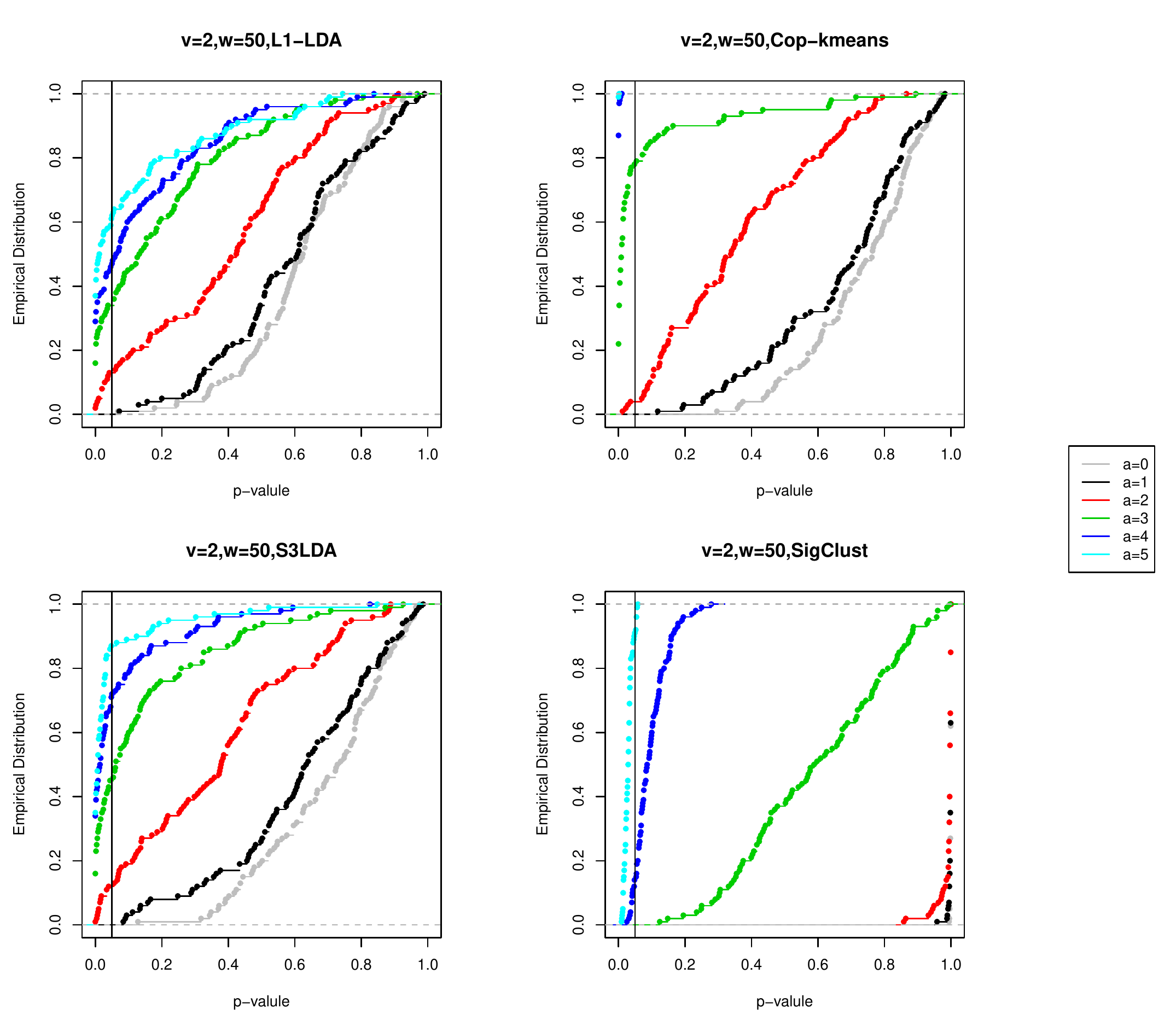}
	\end{center}\vspace{-2ex}
	\caption{Empirical distributions of $p$-values of a mixture of two Gaussian distributions with the signal in one direction. Results are based on different methods under the setting $v=2$ and $w=50$, with the increase of signal $a$.}
	\label{fig:power_1}
\end{figure}

\begin{figure}[!ht]\vspace{-1.5ex}
	\begin{center}
	\includegraphics[height=0.8\linewidth]{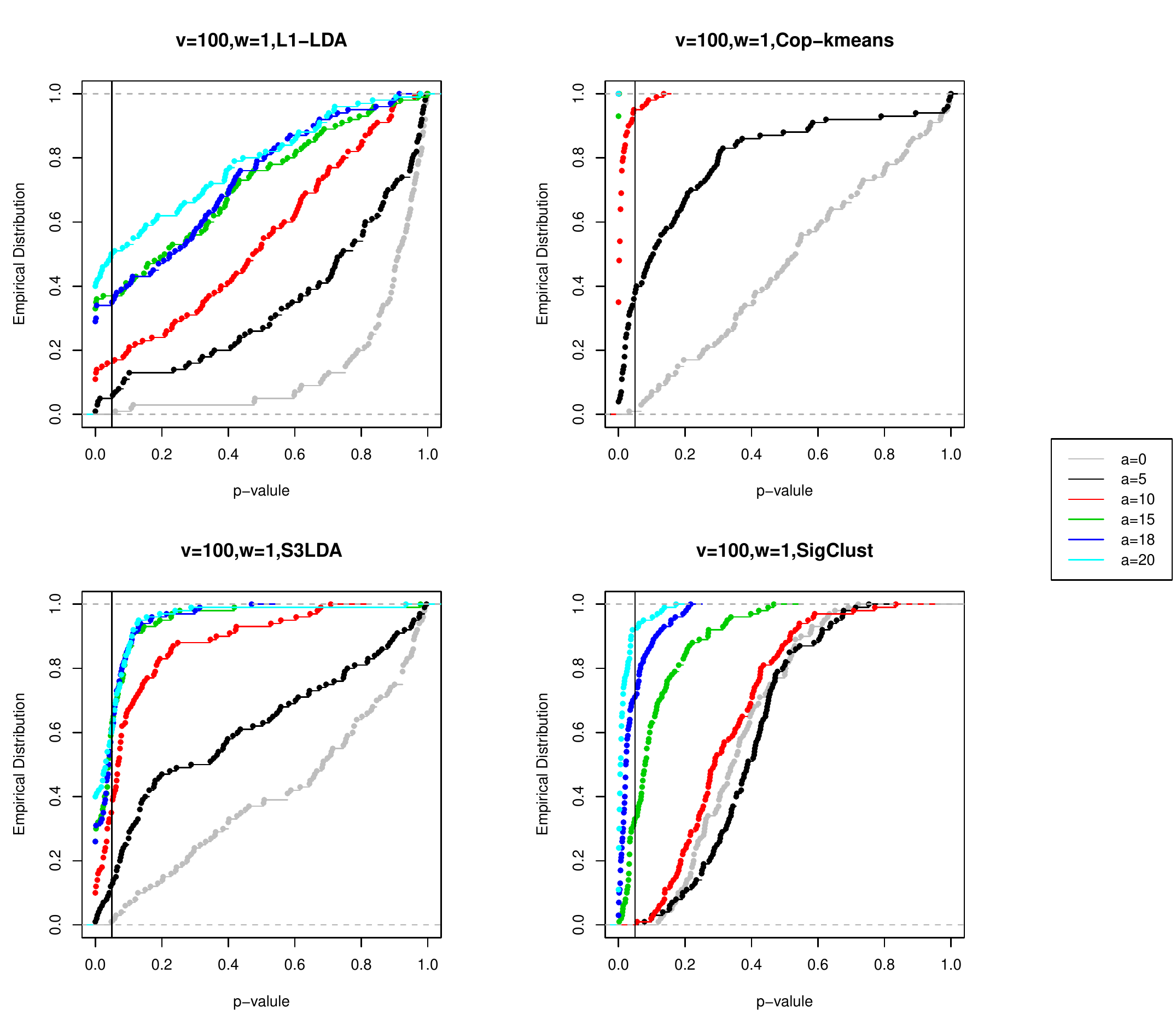}
	\end{center}\vspace{-2ex}
	\caption{Empirical distributions of $p$-values of a mixture of two Gaussian distributions with the signal in one direction. Results are based on different methods under the setting $v=100$ and $w=1$, with the increase of signal $a$.}
	\label{fig:power_2}
\end{figure}

We now consider data generated from a mixture of two Gaussian distributions, $0.5N(-\muv,\Dbf)+0.5N(\muv,\Dbf)$, where $\muv=(a,0,\cdots,0)'$ and $\Dbf=\diag(\underbrace{v,\cdots,v}_{w},0,\cdots,0)$ a diagonal matrix. The sample size is $n=40$. From each class, we randomly choose 10 observations which we keep class labels for. Two types of the covariance matrix are conducted here, $v=100$, $w=1$ and $v=2$, $w=50$. The choices of $a$ depend on the values of $v$ and $w$. Note that when $a=0$, the distribution reduces to a single Gaussian distribution. When $a>0$, the population is a mixture of two Gaussian distributions and the larger the $a$, the greater the separation between the two distributions is. When the signal $a$ is large enough, labeled data do not help on distinguishing the two distributions (they may even make it worse.) Thus one may ignore the label information and simply apply SigClust. In our study, we focus on the cases with small signals, in other words, when the mean difference of the two distributions is not too large. In these cases, labeled data can greatly help to gain extra power in SigPal. The empirical distributions of $p$-values are shown in Figure \ref{fig:power_1} and \ref{fig:power_2} for the two settings ($v=100$, $w=1$ and $v=2$, $w=50$).

%
%
%

Figure \ref{fig:power_1} shows the setting $v=2$ and $w=50$ and Figure \ref{fig:power_2} shows the spiked model setting $v=100$ and $w=1$. Colors are used to represent different values of $a$. When $a=0$, the data are generated from a single Gaussian distribution. When $a>0$, we study the power of the test using different methods. We consider $a=1,\cdots,5$ for the setting $v=2$ and $w=50$ and $a=5,10,15,18,20$ for the setting $v=100$ and $w=1$. We can see in Figure \ref{fig:power_1} that SigClust is too conservative when $a=1,2,3$ and there is almost no power when $a=1,2$. All the three SigPal methods present more power in these settings.

For the spiked model setting in Figure \ref{fig:power_2} where $v=100$ and $w=1$, SigClust is anti-conservative, indicated by the fact that the $p$-value has a higher chance of having a smaller value (the grey curve is above the 45 degree line.) On the other hand, $S^3$LDA and COP-KMEANS are more powerful than SigClust. $L_1$-LDA loses some power as it only uses the labeled data on the simulated data assignments. The comparison on the left two subfigures ($L_1$-LDA versus $S^3$LDA) also illustrates the effect of using a semi-supervised classification for label assignment compared to using a classification method. A greater power is retained as a consequence.

\begin{figure}[!b]\vspace{-1.5ex}
	\begin{center}
	\includegraphics[height=0.8\linewidth]{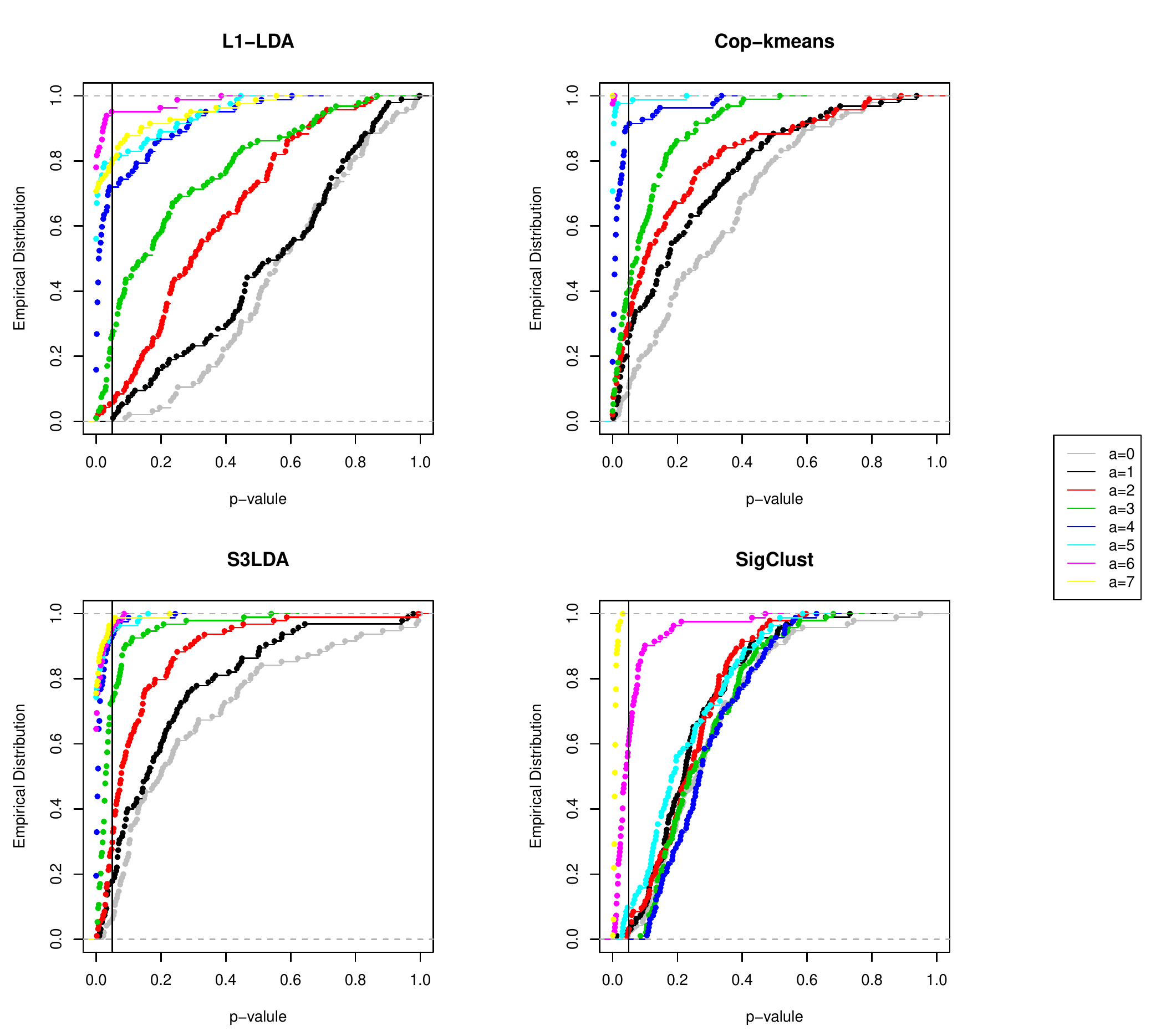}
	\end{center}\vspace{-2ex}
	\caption{Empirical distributions of $p$-values of a mixture of two Gaussian distributions, generated by the covariance matrix from the real data, with the signal in one direction.}
	\label{fig:power_3_D2}
\end{figure}

To make the simulation closer to the reality, we also use the Human Lung Carcinomas Microarray Dataset to obtain a more realistic covariance structure $\Dbf$. This data set was previously analyzed in \citet{bhattacharjee2001classification}. \citet{liu2008statistical} used this data as a test bed to demonstrate their proposed SigClust approach. We extract four biological groups in the data set, 20 pulmonary carcinoid samples (Carcinoid), 13 colon cancer metastasis samples (Colon), 17 normal lung samples (normal) and 6 small cell carcinoma samples (SmallCell), with a total of 56 samples. We remove the first three principal components of the data by reconstructing the covariance matrix using the remaining terms of the eigen-expansion. The resulting covariance $\Dbf$ is used to generate the original data, and the estimated covariance is used to generate the simulated data in the \textbf{Simulation} step of SigPal.

In terms of the signal, we consider the signal $a$ on one direction ranging from $1,2,\cdots,7$. The empirical distributions of $p$-values are displayed in Figure \ref{fig:power_3_D2}. 


Figure \ref{fig:power_3_D2} shows that all of $S^3$LDA, COP-KMEANS and SigClust are strongly anti-conservative under the null hypothesis ($a=0$), which is not the case in either Figure \ref{fig:power_1} or Figure \ref{fig:power_2}. The data in Figure \ref{fig:power_3_D2} is generated from a non-diagonal covariance matrix $\Dbf$ while Figure \ref{fig:power_1} and \ref{fig:power_2} use a diagonal covariance $\Dbf$. Since CI is rotation invariant, it suggests that these methods become anti-conservative due to the estimation of covariance matrix.

\begin{figure}[!ht]\vspace{-1.5ex}
	\begin{center}
	\includegraphics[height=0.8\linewidth]{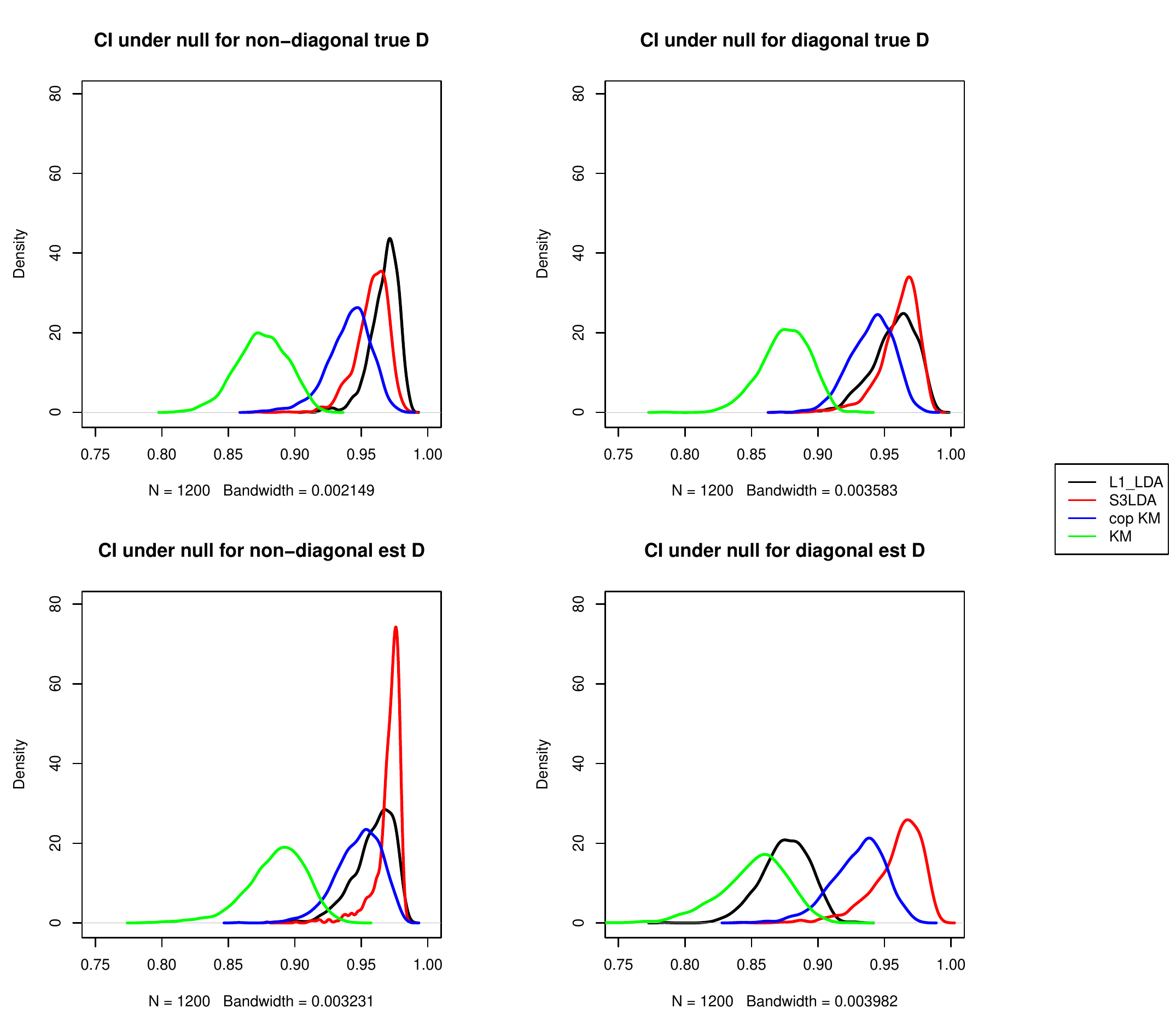}
	\end{center}\vspace{-2ex}
	\caption{Understanding why our method is anti-conservative for non-diagonal covariance by comparing with the diagonal case.}
	\label{fig:power_3_understanding}
\end{figure}

To further understand the influence from the estimation of diagonal and non-diagonal covariances, we compare the distributions of CI under null hypothesis in Figure \ref{fig:power_3_understanding}. The left two plots show the distributions of CI for the data generated from a non-diagonal covariance matrix (top) and for the data generated from the estimated covariance (bottom). It shows that the density curves of CI for the simulated data (bottom) for $S^3$LDA, COP-KMEANS and SigClust (red, blue and green curves) are all shifted to the right compared to the case with the true distribution (top). The CI for the simulated data (bottom) is more likely to be greater than the CI for the original data (top), which makes the $p$-value small more often and hence the three methods become anti-conservative. This is consistent with the result we see in Figure \ref{fig:power_3_D2}.

\begin{figure}[!b]\vspace{-1.5ex}
	\begin{center}
	\includegraphics[height=0.8\linewidth]{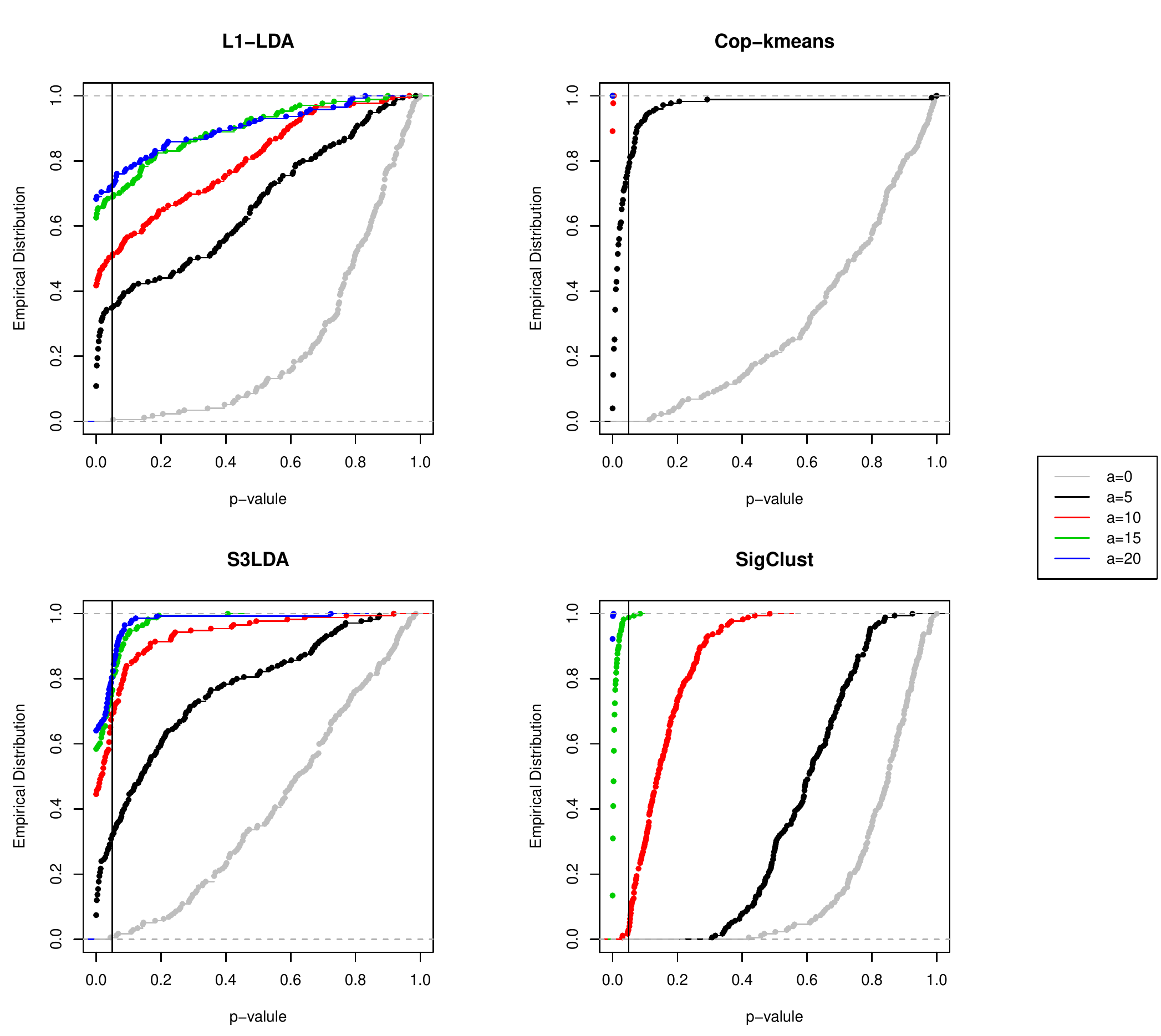}
	\end{center}\vspace{-2ex}
	\caption{Empirical distributions of $p$-values of a mixture of two Gaussian distributions, generated by a diagonal covariance matrix from the real data, with the signal in one direction.}
	\label{fig:power_3_D1}
\end{figure}

For the right two plots, we rotate and obtain a diagonal covariance matrix by eigen-decomposition and plot the empirical distributions of CI for the data generated from the true diagonal covariance (top) and from its estimation (bottom). The density curves of the CI for the simulated data (bottom) for $S^3$LDA, COP-KMEANS and SigClust almost remain in the same position as those using the true covariance; the curve for $L_1$-LDA shifts greatly to the left. Based on the comparison between the left and the right panels, we confirm our previous finding that $S^3$LDA, COP-KMEANS and SigClust become anti-conservative when $\Dbf$ is non-diagonal due to the influence from covariance estimation.

After conducting eigen-decomposition of the covariance matrix used in Figure \ref{fig:power_3_D2}, we obtain a diagonal $\Dbf$ and use it to generate the data. New results are presented in Figure \ref{fig:power_3_D1}. None of the methods is anti-conservative and all the three versions of SigPal are more powerful than SigClust when the signal $a$ is relatively small ($a=5,10$). One is recommended to rotate the data to form a diagonal covariance matrix before applying SigPal.

\subsubsection{Case 3: Mixture of Two Gaussian Distributions With Signal in All Coordinate Directions}

\begin{figure}[!b]\vspace{-1.5ex}
	\begin{center}
	\includegraphics[height=0.8\linewidth]{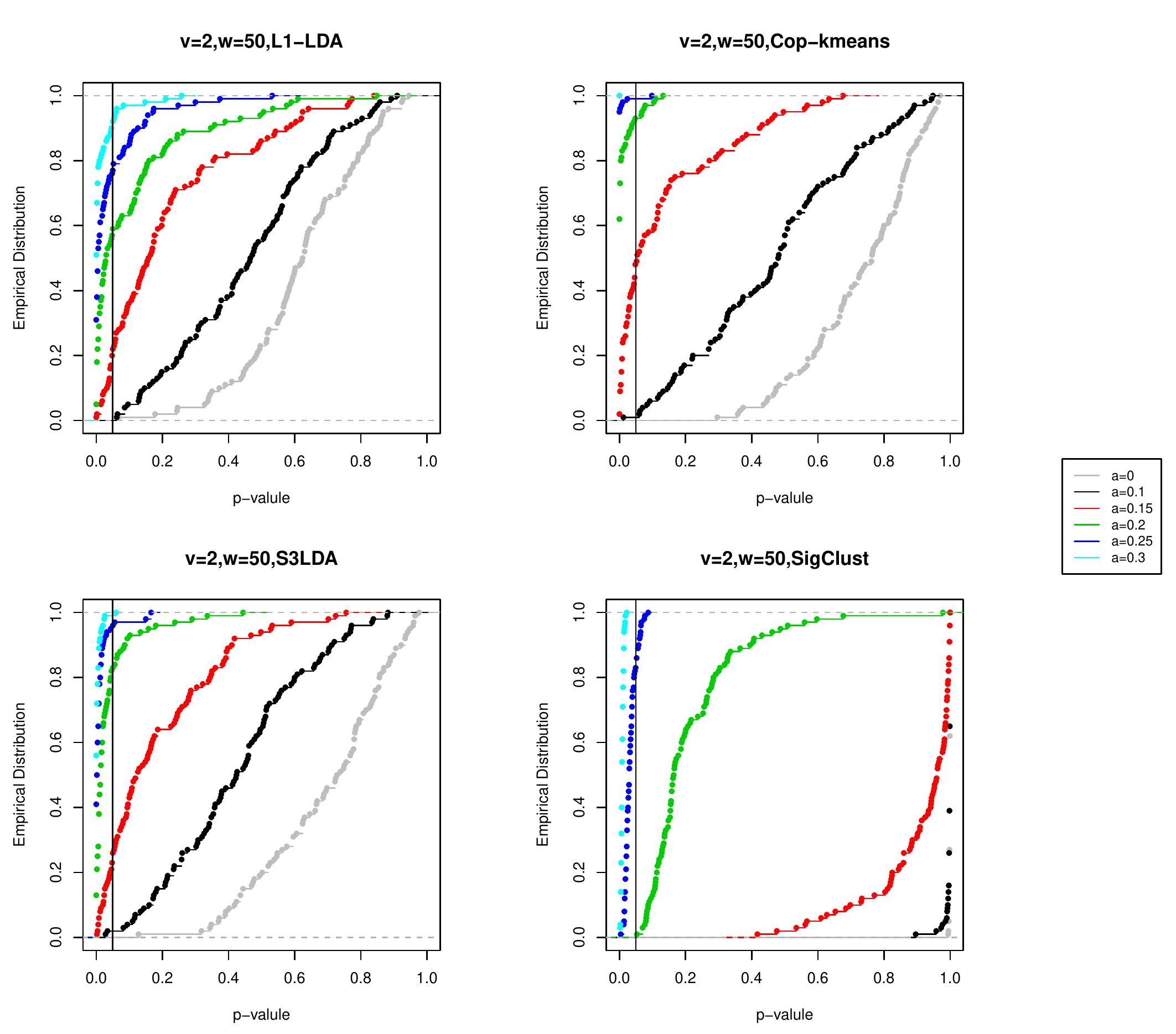}
	\end{center}\vspace{-2ex}
	\caption{Empirical distributions of $p$-values of a mixture of two Gaussian distributions with the signal in all directions. Results are based on different methods under the setting $v=2$ and $w=50$, with the increase of signal $a$.}
	\label{fig:power_4}
\end{figure}

\begin{figure}[!b]\vspace{-1.5ex}
	\begin{center}
	\includegraphics[height=0.8\linewidth]{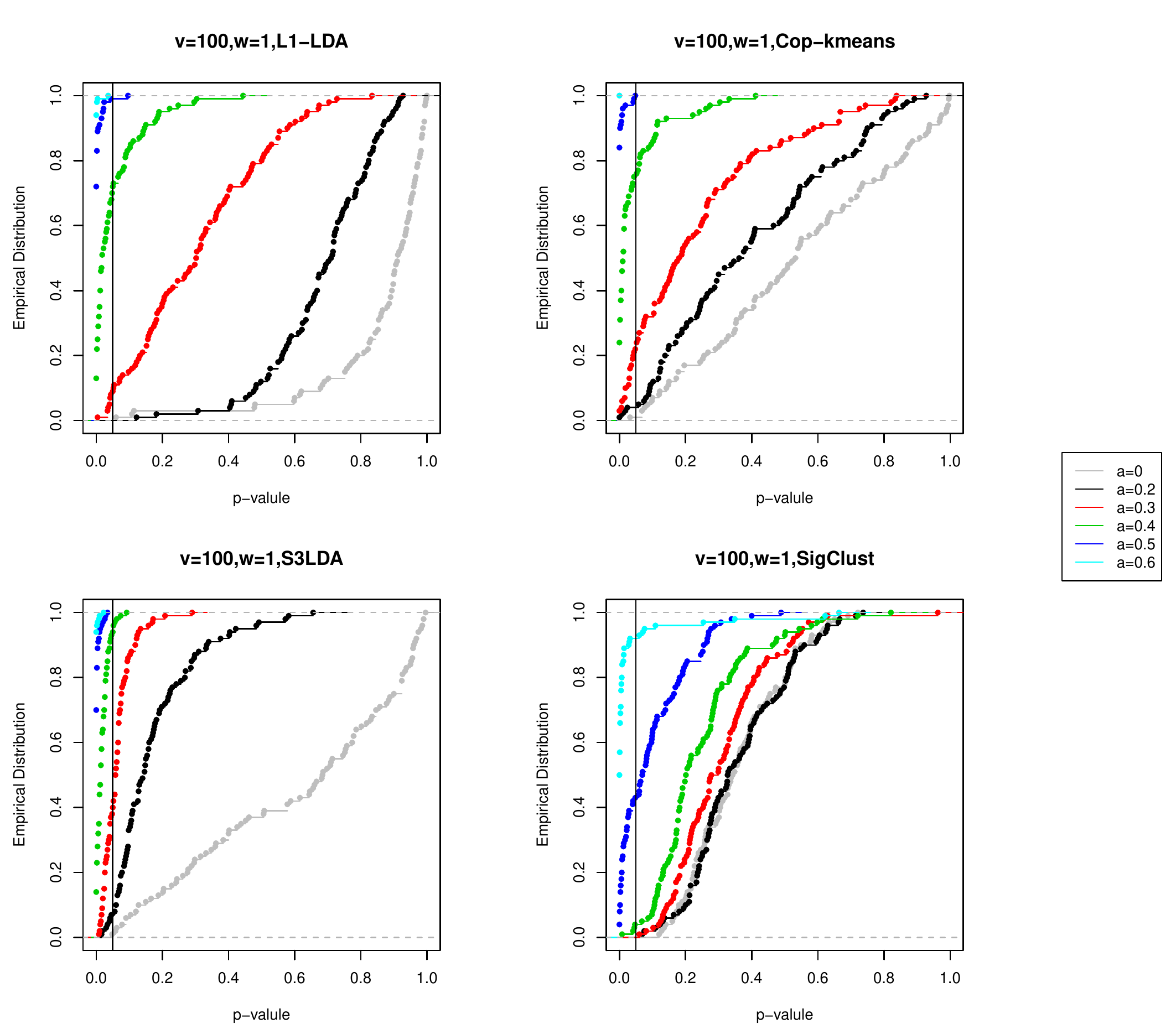}
	\end{center}\vspace{-2ex}
	\caption{Empirical distributions of $p$-values of a mixture of two Gaussian distributions with the signal in all directions. Results are based on different methods under the setting $v=100$ and $w=1$, with the increase of signal $a$.}
	\label{fig:power_5}
\end{figure}

Similarly as in Figure \ref{fig:power_1}, we see in Figure \ref{fig:power_4} that SigClust is too conservative when $a=0.1$ and 0.15. All the three SigPal methods perform more powerfully than SigClust when $a$ is less than 0.25. For the spiked model in Figure \ref{fig:power_5} where $v=100$ and $w=1$, SigClust is anti-conservative. SigPal is more powerful than SigClust when $a\leq 0.6$.

Now we further consider examples with signals in all coordinate directions. We generate data from a mixture of two Gaussian distributions, $0.5N(-\muv,\Dbf)+0.5N(\muv,\Dbf)$, where $\muv=(a,\cdots,a)'$ and $\Dbf=\diag(\underbrace{v,\cdots,v}_{w},0,\cdots,0)$ a diagonal matrix. We keep the class labels for 10 observations from each class and still consider two covariance settings, $v=100$, $w=1$ and $v=2$, $w=50$. The signal $a$ in each direction is deliberately chosen to be very small, however, when all directions are combined together, the total signal can be very large. The empirical distributions of $p$-values calculated from 100 replications for the two settings are displayed in Figure \ref{fig:power_4} and \ref{fig:power_5}.

%

In summary, Sigpal maintains the size under the null distribution while SigClust is anti-conservative when the first eigenvalue is very large relative to the others. In all the cases when the signal between the two distributions is small, SigPal is relatively more powerful than SigClust due to the help from labeled data. Among the three versions of SigPal we consider in the simulation study, COP-KMEANS performs the best in most cases. When the data follows a distribution with non-diagonal covariance matrix, the test could be anti-conservative. Thus rotation of the data is recommended before SigPal is applied.

\subsection{Real Data Application}

In this section, we apply our method to the breast cancer data (BRCA) from The Cancer Genome Atlas Research Network, which has been studied by \citet{fan2006concordance} and \citet{huang2014statistical}. The data include four subtypes: LumA, LumB, Her2 and Basal. The sample size is 348, among which there are 154 LumA, 81 LumB, 42 Her2 and 66 Basal. The number of genes used in the analysis is 4000 after filtering. For every possible pairwise combination of subclasses, we randomly select 20 observations from each class to keep the class labels and the remaining observations are treated as unlabeled. We apply SigClust, SigPal and DiProPerm to every possible pair of subclasses and report their $p$-values in Table \ref{realdata}. Here we only conduct SigPal using COP-KMEANS for class assignment in this real example. Note that these three methods are  using different information. SigClust does not require label information while DiProPerm is applied to the two labeled classes. Our SigPal method is designed for partially labeled data.

\begin{table}[!t]\footnotesize
\centering
\begin{tabular}{ccccccc}
\hline
           &    Basal.LumA & Basal.LumB & Basal.Her2     & LumA.LumB & Her2.LumB  &Her2.LumA  \\\hline
 SigClust  &             0 &       0    &      0.009		 &    0.298  &    0.537   &   0.625   \\
 SigPal    &             0 &       0    &        0		   &    0.002  &    0.002   &   0.013   \\
 DiProPerm &             0 &       0    &        0       &        0  &        0   &       0   \\\hline
\end{tabular}
\caption{SigClust, SigPal and DiProPerm $p$-values for each pair of subtypes for the BRCA data. With the label information, DiProPerm can always reach significant results for all pairs. With only partial information, SigPal can reach similar conclusions. }\label{realdata}
\end{table}

Table \ref{realdata} shows that for pairs including Basal, the $p$-values from all three methods are 0 which implies that Basal can be well separated from the rest. For the remaining three pairs, SigClust reports large $p$-values, which suggests that there is no strong evidence for them to be viewed as from two different clusters if no label information is provided. In contrast, the $p$-values of these three pairs for DiProPerm are all 0, indicating that each pair of two classes can be significantly separated. With the help of a small portion of the label information, SigPal gains much power to distinguish the two classes and produces very close results to DiProPerm.

\begin{figure}[!b]\vspace{-1.5ex}
	\begin{center}
	\includegraphics[height=0.4\linewidth]{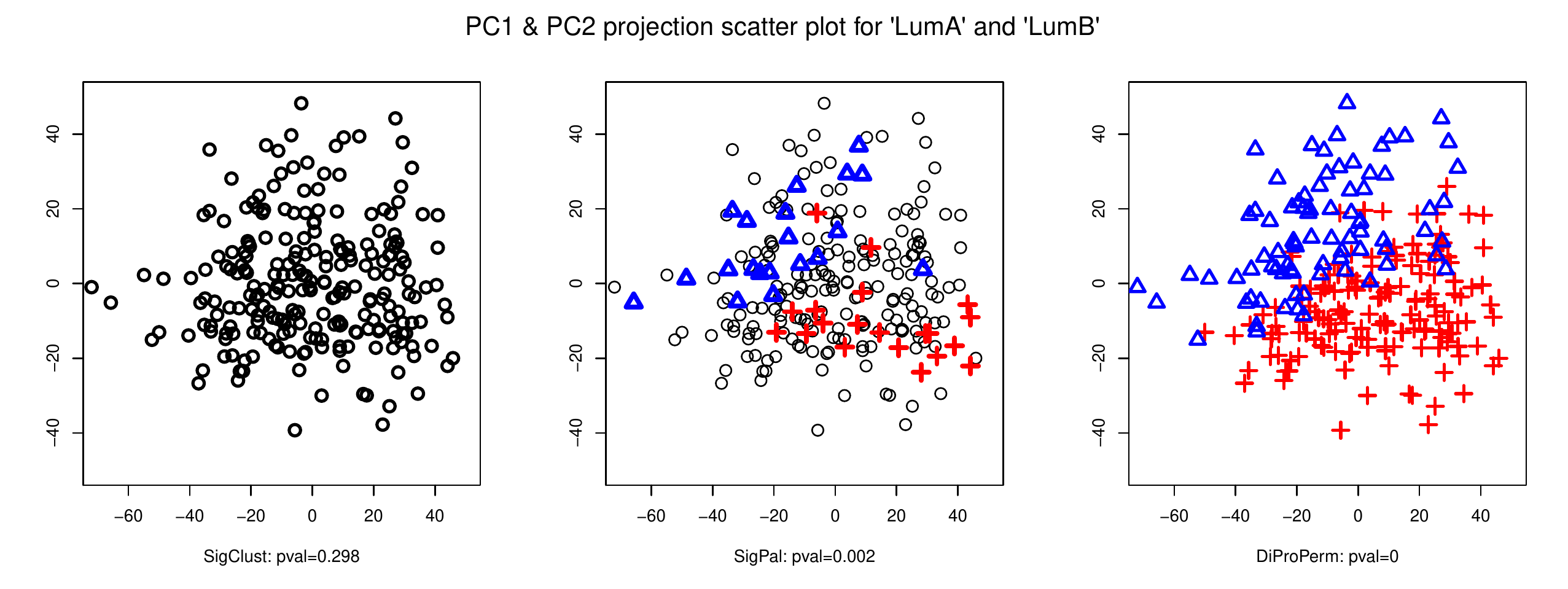}
	\end{center}\vspace{-2ex}
	\caption{PCA projection scatter plot of the two classes, LumA and LumB. Colors indicate biological subtypes. LumA are displayed in red and LumB in Blue. Points in black are treated as unlabeled data. Data are analyzed by SigClust, SigPal and DiProPerm respectively in the left, middle and right panels.}
	\label{fig:realdatapca}
\end{figure}

To further illustrate the three methods, we use the pair of LumA and LumB as an example and display the scatter plot of the projections of the data vectors onto the first two principle component directions in Figure \ref{fig:realdatapca}. Colors indicate biological subtypes. LumA are displayed in red and LumB in Blue. Points in black are treated as unlabeled data.  Figure \ref{fig:realdatapca} shows that without given the class information (left plot), LumA and LumB seem to be one subtype so that SigClust give a non-significant result ($p$-value=0.298). When all the label information is available (right plot), DiProPerm suggests that these two classes are significantly different ($p$-value=0). With 40 labeled observation out of 235 observations in total, our SigPal method finds the difference between the two classes by extracting useful information from the small portion of the labeled data (middle plot).

\section{Conclusion}\label{conclusion}

In this article, we propose a significance analysis procedure, SigPal, in the HDLSS setting. This method is designed for a data set where a small amount of labeled data are available with a large amount of unlabeled data. In contrast to SigClust which does not rely on class label information, our method makes use of the labeled data to increase the difference in the classes under the null and alternative hypotheses. Through extensive simulation examples with partial label information available, we compared the performance of SigPal with SigClust in different settings. SigPal is relatively more powerful than SigClust, especially when the signal between the two classes is not large. Among the three versions of SigPal we conduct in the simulation study, COP-KMEANS performs the best in most cases. 

Although CI is rotation invariant, it turns out in the simulation study that under the null hypothesis when the data comes from a distribution with non-diagonal covariance matrix, SigPal could be anti-conservative. Hence, rotation of the data is recommended before SigPal is applied. 

SigPal is a general procedure with possibly many variants. The test statistic CI, used in our numerical study, may be substituted by other quantities, such as the Hotelling's $T^2$ statistic. There is also room for choosing different approaches to assign labels in the \textbf{Initialization} step of SigPal. An interesting and potential extension of SigPal is to case of testing multiple classes. A possible solution is to use a multi-class classification method for the class assignment. 


\section*{Appendix. Technical proofs}
\subsection*{Proof of Theorem~\ref{theorem_tci}}

Without loss of generality, assume that $\Sigma=\diag(\lambda_1,\cdots,\lambda_d)$ with $\lambda_1\geq\lambda_2\geq\cdots\geq\lambda_d$. We first show that $\hat{\omegav}=\vv_1=(1,0,\cdots,0)'$, which is the direction of the greatest variation of the data. Recall that $\hat{\omegav}=\argmin_{\|\omegav\|=1}~\E_{(\Xv,Y)}(Y-\omegav'\Xv)^2+C\E_{\Xv}(1-|\omegav'\Xv|)_+$. As $C\rightarrow\infty$, we only need to show that $\vv_1=(1,0,\cdots,0)'$ minimizes the second term 
\begin{align*}
  & \E[(1-|Z|)_+] \\
= & \Pr(|Z|\geq 1)\E(0\big{|}|Z|\geq 1)+\Pr(|Z|<1)\E(1-|Z|\big{|}|Z|<1) \\
= & 0+\Pr(|Z|<1)\E(1-|Z|\big{|}|Z|<1),
\end{align*}
where $Z=\omegav'\Xv$.

Let $\omegav_1=(1,0,\cdots,0)'$ and $\omegav_2=(s_1,\cdots,s_d)'$ with $\sum_{j=1}^d s_j^2=1$. Then we have $\var(\omegav_1'\Xv)=\omegav_1'\Sigma\omegav_1=\lambda_1\geq\var(\omegav_2'\Xv)$. Let $Z_1=\omegav_1'\Xv$ and $Z_2=\omegav_2'\Xv$, then $Z_1$ and $Z_2$ follow Gaussian distributions with mean 0 and $\var(Z_1)\geq \var(Z_2)$. It follows that 

1. $\Pr(|Z_1|<1)\leq\Pr(|Z_2|<1)$;

2. $\E(|Z_1|\big{|}|Z_1|<1)\geq\E(|Z_2|\big{|}|Z_2|<1)$. 

Therefore, $\E[(1-|Z_1|)_+]\leq\E[(1-|Z_2|)_+]$, which implies that $\hat{\omegav}=(1,0,\cdots,0)'$.

To show the TCI, it is enough to consider the situation with diagonal covariance matrix due to the rotation invariance of CI. We first compute the theoretical total sum of squares TSS as
\begin{align*} 
TSS & = \E\|\Xv\|^2=\int\|\xv\|^2\phi(\xv)d\xv \\
    & = \int_{-\infty}^{\infty}\cdots\int_{-\infty}^{\infty}\|\xv\|^2\phi(\xv)dx_1\cdots dx_d \\
		& = \int_{-\infty}^{\infty}\cdots\int_{-\infty}^{\infty}\sum_{j=1}^d x_j^2\Big(\prod_{j=1}^d \varphi_{\lambda_j}(x_j)\Big)dx_1\cdots dx_d \\
		& = \sum_{j=1}^d\int_{-\infty}^{\infty}x_j^2 \varphi_{\lambda_j}(x_j)dx_j = \sum_{j=1}^d\lambda_j,
\end{align*}
where $\phi(\xv)=\prod_{j=1}^d\varphi_{\lambda_j}(x_j)=\prod_{j=1}^d\frac{1}{\sqrt{2\pi\lambda_j}}e^{-x_j^2/2\lambda_j}$. 

Recall that we assume $\lambda_1\geq\lambda_2\geq\cdots\geq\lambda_d$ and we showed $\hat\omegav=\vv_1=(1,0,\cdots,0)'$. Here $(1,0,\cdots,0)'$ is the norm vector of the separating hyperplane going through $\muv=(0,\cdots,0)'$. 

Let WSS be the theoretical within cluster sum of square and let $\text{WSS}_1$ and $\text{WSS}_2$ denote the theoretical sum of squares within class 1 and class 2 respectively. By symmetry the mean of class 1 $\muv_1=(\mu_{11},\mu_{12},\cdots,\mu_{1d})'$ with $\mu_{12}=\mu_{13}=\cdots=\mu_{1d}=0$. For the first dimension, note that class 1 contains the original labeled data with mean 0 with probability $\theta$, and the original unlabeled data assigned to class 1 with mean $2\int_0^{\infty}x_1 \varphi_{\lambda_1}(x_1)dx_1$ with probability $(1-\theta)$, where $\theta$ is the proportion of the labeled data. Thus we have 
$$\mu_{11}=(1-\theta)\cdot 2\int_0^{\infty}x_1 \varphi_{\lambda_1}(x_1)dx_1+\theta\cdot 0=(1-\theta)\sqrt{\frac{2\lambda_1}{\pi}}.$$ 
So $\muv_1=((1-\theta)\sqrt{\frac{2\lambda_1}{\pi}},0,\cdots,0')$. Similarly, $\muv_2=(-(1-\theta)\sqrt{\frac{2\lambda_1}{\pi}},0,\cdots,0)'$. Then 
\begin{align*}
WSS_1 = & ~(1-\theta)\int_{0}^{\infty}\cdots\int_{-\infty}^{\infty}\|\xv-\muv_1\|^2\phi(\xv)dx_1\cdots dx_d  \\
				&	\quad +\theta\int_{-\infty}^{\infty}\cdots\int_{-\infty}^{\infty}\|\xv-\muv_1\|^2\phi(\xv)dx_1\cdots dx_d \\
			= & ~(1-\theta)\int_{0}^{\infty}\Big(x_1-(1-\theta)\sqrt{\frac{2\lambda_1}{\pi}}\Big)^2\varphi_{\lambda_1}(x_1)dx_1 
					+(1-\theta)\sum_{j=2}^d\int_{0}^{\infty}\int_{-\infty}^{\infty}\cdots\int_{-\infty}^{\infty}x_j^2\phi(\xv)dx_1\cdots dx_d \\
				& ~\quad +\theta\int_{-\infty}^{\infty}\Big(x_1-(1-\theta)\sqrt{\frac{2\lambda_1}{\pi}}\Big)^2\varphi_{\lambda_1}(x_1)dx_1 
				  +\theta\sum_{j=2}^d\int_{-\infty}^{\infty}\int_{-\infty}^{\infty}\cdots\int_{-\infty}^{\infty}x_j^2\phi(\xv)dx_1\cdots dx_d   \\ 
			=	& ~\left[\half(\theta+1)+\frac{1}{\pi}(\theta^3-3\theta^2+3\theta-1)\right]\lambda_1+\sum_{j=2}^d\frac{1+\theta}{2}\lambda_j.
\end{align*}
Similarly, $WSS_2=WSS_1$. Thus,
$$TCI=\frac{WSS_1+WSS_2}{TSS}=1+\theta-\frac{2}{\pi}(1-\theta)^3\frac{\lambda_1}{\sum_{j=1}^d\lambda_j}.\qed$$

\subsection*{Proof of Theorem~\ref{theorem_asymp}}

The proof is similar to the proof of Theorem 1 in \citet{liu2008statistical}. It is sufficient to show the following two points:

1. The CI $\xi_1$ of the data from the mixture of two Gaussian distributions, using the sources of each observation from the two Gaussian distributions as cluster assignments, converges to 0 in probability as $d\rightarrow\infty$.

2. The CI under the null hypothesis is bounded away from 0 as $d\rightarrow\infty$.

Point 1 can be shown by introducing a new data set, which is easier to work with, and a modified corresponding CI, $\xi_2$. In particular, consider iid sample $\yv_1,\cdots,\yv_n$ from $N(\0v,\Dbf)$. Note that $\xv_i\stackrel{d}{=}\yv_i+\delta\muv$, where $\delta=0$ if $\xv_i$ comes from $N(\0v,\Dbf)$ and 1 if $\xv_i$ comes from $N(\muv,\Dbf)$. $\xv_i\stackrel{d}{=}\yv_i+\delta\muv$ implies that $\xi_1\stackrel{d}{=}\xi_2$. Let $C_{(1)}$ and $C_{(2)}$ denote the sample index sets of $\xv_i$ with $\delta=0$ and $\delta=1$ respectively. By definition, we have
$$\xi_1=\frac{\sum_{k=1}^2\sum_{i\in C_{(k)}}\|\xv_i-\bar\xv^{(k)}\|^2}{\sum_{i=1}^n\|\xv_i-\bar\xv\|^2}.$$
$\xi_2$ can be written using $\yv_1,\cdots,\yv_n$, $a_1,\cdots,a_d$ and $d$ as 
\begin{align}\label{eq:point_1}
\xi_2 & = \frac{\sum_{k=1}^2\sum_{i\in C_k}\|\yv_i-\bar\yv^{(k)}\|^2}{\sum_{i=1}^n\|\yv_i-\bar\yv\|^2+\frac{n_1n_2}{n}\sum_{j=1}^d a_j^2+\frac{2n_1n_2}{n}\sum_{j=1}^d a_j(\bar y_j^{(1)}-\bar y_j^{(2)})}  \nonumber\\
      & \leq \frac{\sum_{i=1}^n\|\yv_i-\bar\yv\|^2}{\frac{n_1n_2}{n}\sum_{j=1}^d a_j^2+\frac{2n_1n_2}{n}\sum_{j=1}^d a_j(\bar y_j^{(1)}-\bar y_j^{(2)})}  \nonumber\\
			& =\frac{ d^{-1}\sum_{j=1}^d\sum_{i=1}^n(y_{ij}-\bar y_j)^2}{\frac{n_1n_2}{n}\sum_{j=1}^d a_j^2d^{-1}+\frac{2n_1n_2}{n}\sum_{j=1}^d a_j(\bar y_j^{(1)}-\bar y_j^{(2)})d^{-1}},  
\end{align}
where $C_1$ and $C_2$ denote the random grouping indices of the sample into two classes of size $n_1$ and $n_2$, $\bar y_j$ and $\bar y_j^{(k)}$ denote the overall sample mean and the sample mean of class $k$ of the $j$th variable. Note that $\sum_{i=1}^n(y_{ij}-\bar y_j)^2\sim\lambda_j\chi^2(n-1)$, $\sum_{j=1}^d\lambda_j=O(d^{\beta})$ and $\beta<1$. Thus both the mean and variance of the numerator in (\ref{eq:point_1}) converge to 0 as $d\rightarrow \infty$, which implies that the numerator of (\ref{eq:point_1}) converges to 0 in probability.

For the denominator of (\ref{eq:point_1}), since $\sum_{j=1}^d a_j^2=O(d)$, the first term converges to a constant as $d\rightarrow \infty$. For the second term, note that $\sum_{j=1}^d a_j(\bar y_j^{(1)}-\bar y_j^{(2)})\sim N(0,\sum_{j=1}^d a_j^2\lambda_j(\frac{1}{n_1}+\frac{1}{n_2}))$. Because $\sum_{j=1}^d a_j^2\lambda_j=O(d^{\gamma})$ with $\gamma< 2$, the second term of the denominator of (\ref{eq:point_1}) converges to 0 in probability as $d\rightarrow \infty$. Therefore, $\xi_2\rightarrow 0$ in probability as $d\rightarrow \infty$, which implies $\xi_1\rightarrow 0$ in probability as $d\rightarrow \infty$.

To show point 2, We first get the Gaussian null distribution of the mixture as $N(\0v,\Dbf^*)$, where $\Dbf^*$ is diagonal with the $j$th diagonal element $\lambda_j+\eta(1-\eta)a_j^2$. Here we simply assume the null distribution with mean $\0v$ since CI is location invariant. Let $\zv_1,\cdots,\zv_n$ be a sample from the Gaussian null distribution. We want to show that the corresponding CI is bounded away from 0 as $d\rightarrow \infty$. To this end, we make use of the HDLSS geometry of \citet{Hall2005Geometric}. We first check the three assumptions:

(1) the fourth moments of all entries of $\zv$ are uniformly bounded, by the assumption that $\max_j(\lambda_j+\eta(1-\eta)a_j^2)\leq M$;

(2) $\lim_{d\rightarrow\infty}\text{trace}(\Dbf^*)/d=\sigma^2$, where $\sigma^2$ is a constant. This is because $\lim_{d\rightarrow\infty}\text{trace}(\Dbf^*)/d=\lim_{d\rightarrow\infty}\sum_{j=1}^d(\lambda_j+\eta(1-\eta)a_j^2)/d=\eta(1-\eta)a^2\equiv\sigma^2$, where $a^2=\lim_{d\rightarrow\infty}\sum_{j=1}^d a_j^2/d$;

(3) the random vector satisfies the $\rho$-mixing condition by the independence among the entries of $\zv$.

Then it follows that $\|\zv_i-\zv_l\|^2=2d\sigma^2+O_p(1)$, as $d\rightarrow \infty$. By the triangle inequality, we have $2(\|\zv_i-\bar\zv\|^2+\|\zv_l-\bar\zv\|^2)\geq (\|\zv_i-\bar\zv\|+\|\zv_l-\bar\zv\|)^2\geq \|\zv_i-\zv_l\|^2$ and $\|\zv_i-\bar\zv\|=\frac{1}{n}\Big\|(n-1)\zv_i-\sum_{l\ne i}\zv_l\Big\|\leq \frac{1}{n}\sum_{l\ne i}\|\zv_i-\zv_l\|$. Thus, as $d\rightarrow \infty$, we can bound the CI under the null hypothesis as 
\begin{align*}
CI & = \frac{\sum_{k=1}^2\sum_{i\in C_{k}}\|\zv_i-\bar\zv^{k}\|^2}{\sum_{i=1}^n\|\zv_i-\bar\zv\|^2} \\
	 & \geq \frac{\frac{1}{2}([n_1/2]+[n_2/2])2d\sigma^2+O_p(1)}{\frac{(n-1)^2}{n}2d\sigma^2+O_p(1)} \\
	 & = \frac{n([n_1/2]+[n_2/2])}{2(n-1)^2}+o_p(1),
\end{align*}
where $[u]$ denotes the largest integer smaller than $u$. This shows that under the null hypothesis CI is bounded away from 0 as $d\rightarrow\infty$.\qed

\bibliographystyle{asa}
\bibliography{sigpal-reference}

\end{document}